\documentclass[10pt,journal,compsoc]{IEEEtran}
\ifCLASSOPTIONcompsoc
  \usepackage[nocompress]{cite}
\else
  \usepackage{cite}
\fi
\ifCLASSINFOpdf
\else
\fi

\usepackage[utf8]{inputenc} 
\usepackage[T1]{fontenc}    
\usepackage{hyperref}       
\usepackage{url}            
\usepackage{booktabs}       
\usepackage{amsfonts}       
\usepackage{nicefrac}       
\usepackage{microtype}      
\usepackage{xcolor}         
\usepackage{microtype}
\usepackage{graphicx}
\usepackage{subfigure}
\usepackage{booktabs} 
\usepackage{multirow}
\usepackage{caption}
\usepackage{tabularx}
\usepackage[ruled,vlined,linesnumbered]{algorithm2e}
\usepackage{mathtools}
\usepackage{diagbox}

\usepackage{array}
\usepackage{amsmath}
\DeclareMathOperator*{\argmin}{arg\,min}
\usepackage{amssymb}
\usepackage{amsthm}

\begin{document}
%
\title{Continual Learning on Graphs: Challenges, Solutions, and Opportunities}
%
%
%
%

\author{Xikun~Zhang, Dongjin~Song*,~\IEEEmembership{Member,~IEEE}, and~Dacheng~Tao*,~\IEEEmembership{Fellow,~IEEE}
\IEEEcompsocitemizethanks{
\IEEEcompsocthanksitem 

Mr X. Zhang and Prof D. Tao are with the School of Computer Science, in the Faculty of Engineering, at The University of Sydney, 6 Cleveland St, Darlington, NSW 2008, Australia. \protect Email: xzha0505@uni.sydney.edu.au, dacheng.tao@sydney.edu.au.
\IEEEcompsocthanksitem Dr. D. Song is with the Department of Computer Science and Engineering, University of Connecticut, Storrs, Connecticut, the United States. \protect Email: dongjin.song@uconn.edu.
\\
* indicates corresponding authors.
}
}

\IEEEtitleabstractindextext{%
\begin{abstract}

Continual learning on graph data has recently attracted paramount attention for its aim to resolve the catastrophic forgetting problem on existing tasks while adapting the sequentially updated model to newly emerged graph tasks. While there have been efforts to summarize progress on continual learning research over Euclidean data, \textit{e.g.}, images and texts, a systematic review of progress in continual learning on graphs, \textit{a.k.a}, continual graph learning (CGL) or lifelong graph learning, is still demanding. Graph data are far more complex in terms of data structures and application scenarios, making CGL task settings, model designs, and applications extremely challenging. To bridge the gap, we provide a comprehensive review of existing continual graph learning (CGL) algorithms by elucidating the different task settings and categorizing the existing methods based on their characteristics. We compare the CGL methods with traditional continual learning techniques and analyze the applicability of the traditional continual learning techniques to CGL tasks. Additionally, we review the benchmark works that are crucial to CGL research. Finally, we discuss the remaining challenges and propose several future directions. We will maintain an up-to-date GitHub repository featuring a comprehensive list of CGL algorithms, accessible at 
\url{https://github.com/UConn-DSIS/Survey-of-Continual-Learning-on-Graphs}.
\end{abstract}

\begin{IEEEkeywords}
Graph representation learning, continual learning, graph neural networks, continual graph learning, CGL, lifelong learning, lifelong graph learning, continual graph representation learning.
\end{IEEEkeywords}}

\maketitle

\IEEEdisplaynontitleabstractindextext

%
\IEEEpeerreviewmaketitle

\IEEEraisesectionheading{\section{Introduction}\label{sec:introduction}}

%
%
%
%
\IEEEPARstart{I}{n} the field of graph representation learning, traditional methods typically assume static graphs, \textit{i.e.}, the structure of graphs as well as their attributed nodes and edges remain constant. However, in many real-world applications, including both node-level and graph-level scenarios, graphs may evolve constantly. This means that either new types of nodes and their associated edges may appear or new types of graphs may be collected. In this case, the graph data may exhibit distribution shift, and a desired model is expected to continually accommodate the new distributions without forgetting the previously learnt knowledge. For example, new categories of research papers (graph nodes) and the accompanying citations (graph edges) will constantly emerge in a citation network. To automatically classify the papers, a document classifier is expected to continually adapt to the distribution of new categories while maintaining the learnt knowledge of the previously observed categories at the same time~\cite{zhang2022cglb,liu2021overcoming,9808404}. For drug discovery, new molecule properties and new molecule categories may be encountered intermittently, and a molecule property predictor has to fit its parameters for new patterns without compromising their prediction performance over existing molecule categories or properties~\cite{zhang2022cglb,liu2021overcoming}. 
In such continual learning scenarios, naively adapting the model to new tasks will encounter the catastrophic forgetting problem, \textit{i.e.}, drastic performance drop on existing tasks after the model parameters are adapted for new tasks. 

A naive solution is to retrain the model over all previously observed data whenever a new task emerges. However, this could be   
infeasible either due to the intractable retraining cost or potential privacy/legal regulation issues.
In light of this, Continual Graph Learning (CGL), which aims to continually learn new tasks without forgetting previously learnt knowledge, has recently received increasingly more attention from various areas~\cite{liu2021overcoming,zhang2022cglb,galke2023lifelong,shen2023graph,lombardo2022continual,kou2020disentangle,carta2022catastrophic,mirtaheri2023history,yuan2023continual,febrinanto2023graph,song2018enriching,ren2023incremental,lin2023gated,javed2019meta,chen2021trafficstream,tan2023futures,perini2022learning,gupta2023grafenne,chen2023continual,wang2023pattern,omeliyanenko2023capskg,wang2022facing}. Due to the complexity of graph data, existing CGL methods are highly heterogeneous in terms of the targeted graph types, learning settings, basic techniques, and evaluation metrics. First, existing methods may propose general techniques or specialized techniques for certain application scenarios (\textit{e.g.} knowledge graph, recommender system, etc.). Second, according to the availability of the task identity during testing, CGL methods may adopt different settings including task-incremental learning (task-IL), domain-incremental learning (domain-IL), and class-incremental learning (class-IL). Besides the incremental setting, the graph learning tasks may also focus on different granularity, including node-level tasks and graph-level tasks. Third, the methodologies employed by various existing approaches align with distinct prevailing streams, such as regularization, memory replay, and parameter isolation. Finally, different from standard learning settings, the performance of a CGL model is concerned with different perspectives including the overall performance, the performance decrease (forgetting), the inter-task interfere, \textit{etc.} Therefore, depending on the research target, different works may adopt different metrics for evaluating the models. These orthogonal dimensions of CGL research present complex challenges and significant obstacles for researchers attempting to enter the field.

To this end, in this paper, we provide a systematic review of the existing methods from the aforementioned four different perspectives. In Section~\ref{sec:background}, we briefly introduce the context information of continual learning, CGL, as well as the difference between CGL and other related research areas on dynamic evolving graphs.
In Section~\ref{sec:Problem Setups}, we explain the problem setup of CGL from the perspectives of basic concepts, task sequence construction, task granularity, and different incremental learning scenarios. 
In Section~\ref{sec:Methods}, we first systematically review the related CGL works following the three mainstreams, including the regularization based, memory-replay based, and parameter-isolation based methods, and then analyze the applicability of traditional continual learning techniques. In Section~\ref{sec:benchmark}, we introduce the recently proposed benchmark works that aim to provide a consistent experimental setting and fair platform for comparing different techniques.  
Finally, in Section~\ref{sec:Opportunities, Future Directions and Discussions}, we discuss existing challenges and highlight the promising future directions for CGL research.

\section{Background}\label{sec:background}

\subsection{Continual Learning}
Continual learning (CL)~\cite{van2019three,parisi2019continual,zhang2022cglb,li2017learning,9915459} targets the catastrophic forgetting problem, \textit{i.e.} the phenomenon that a model's performance on previous tasks decreases drastically after learning the subsequent new tasks. During training, the learning can be formulated as training a model consecutively on a sequence of tasks. When learning each task, the model can only access the data of the current task, while access to previous tasks is usually not allowed. In continual learning, the data of different tasks typically exhibit different distributions, \textit{e.g.} images of different classes. Accordingly, after learning a new task, if the model parameters are adapted solely for the new task, the learnt knowledge of the previous tasks may be overwritten, leading to the catastrophic forgetting problem over previous tasks. During testing, the model will be tested on each learnt task. A desired model should be able to maintain its performance on every learnt task. The evaluation of continual learning models is detailed in Section~\ref{sec:evaluation}.

\subsection{Graph Representation Learning}
Graph representation learning~\cite{kipf2016semi,ICML20,velivckovic2017graph,ICLR20a,KDD19a,zhang2020context,KDD18b,lei2020spherical,9764632,9868157} aims to generate qualified representations for nodes, edges, or the entire graphs, which can further be used for downstream tasks like node classification~\cite{kipf2016semi,ECML20}, link prediction~\cite{lu2011link,TOIS2021,cai2021line}, graph classification~\cite{gilmer2017neural,vogelstein2012graph,9980390}, \textit{etc.}. Graph neural networks (GNNs) are currently the prevailing approaches. To obtain either node/edge level or graph level representations, GNNs would first generate node/edge level representations.  The most popular GNNs follow the message passing neural network (MPNN) framework~\cite{gilmer2017neural}, and can be formulated as,
\begin{align}
    &\mathbf{m}_v^{l+1} = \sum_{u\in \mathcal{N}^1(v)} \mathrm{M}_l (\mathbf{h}^l_v, \mathbf{h}^l_u, \mathbf{x}^e_{v,u}; \boldsymbol{\theta}_l^\mathrm{M}),&\\
    &\mathbf{h}^{l+1}_v = \mathrm{U}_l(\mathbf{h}_v^l,\mathbf{m}_v^{l+1}; \boldsymbol{\theta}_l^\mathrm{U}),&
\end{align}
As shown in the formulations above, different from learning on independent data (\textit{e.g.} images), generating representations on graph data requires properly capturing the valuable topological information (\textit{e.g.} through the message passing). Similarly, compared to classic CL on independent data, CGL also has to properly preserve the highly valuable topological information. Besides, the topological connections may cause some continual learning techniques to be inapplicable. For example, to generate the representation of a single node, the message passing based GNNs would require aggregating information from multi-hop neighbors, and the methods based on memory replay, which rely on storing individual data points, become untenable.

\subsection{Differentiation from Other Related Works} 
In addition to CGL efforts aimed at mitigating the issue of forgetting, there exists research at the intersection of graph representation learning and evolving graphs that primarily addresses alternative challenges~\cite{du2023multi,bo2022ego,hedegaard2023continual,zaman2023cmdgat,xhonneux2020continuous,luo2020learning,tang2020graph,das2022graph,das2022learning,das2022online,wu2023continual}, which could be confusing. 
In this subsection, we clarify the crucial difference between CGL and some easily confusing topics such as dynamic graph learning~\cite{galke2021lifelong,wang2020streaming,han2020graph,yu2018netwalk,nguyen2018continuous,zhou2018dynamic,ma2020streaming,feng2020incremental,bielak2022fildne,he2023dynamically} and few-shot graph learning~\cite{zhou2019meta,guo2021few,yao2020graph,tan2022graph,garcia2017few,yao2020graph}. Dynamic graph learning primarily aims to capture the evolving graph structure and maintain up-to-date graph representations, with access to all prior information, rather than tackling the forgetting issue. Conversely, CGL focuses on resolving the forgetting problem, and the data of previous tasks are typically unavailable. An exception is CGL with inter-task edges, which permits aggregation of past task information via inter-task edges during the GNNs' neighborhood aggregation process. Nevertheless, the labels of prior task data remain inaccessible. Few-shot graph learning is designed for rapid model adaptation to new tasks. During training, few-shot learning models can access all tasks simultaneously, which is not the case for CGL \textcolor{black}{(CGL with inter-task edges deviates slightly)}. During evaluation, few-shot learning models are tested on new tasks after initial fine-tuning, whereas CGL models are evaluated on existing tasks without any fine-tuning. 

Finally, this paper differs from an earlier survey~\cite{10026151} on CGL in several key aspects. Febrinanto \textit{et al.}~\cite{10026151} emerged during the nascent stages of Continual Graph Learning (CGL) research, incorporating a limited scope of studies due to the field's infancy. In contrast, our survey, benefiting from over a year of rapid advance in CGL~\cite{hoang2023universal,liu2023cat,zhang2023continual,sun2023self} and the deep understanding of the CGL problem based on our previous technical contributions in this field~\cite{zhang2022hierarchical,zhang2022cglb,zhang2022sparsified,zhang2023ricci,zhang2024topology}, encompasses a comprehensive review of state-of-the-art techniques. 
Our survey not only elucidates technical approaches but also offers a systematic exploration of CGL's problem setup.  In addition, we also examine different methods' applicability across diverse application scenarios, as summarized in Table~\ref{tab:summarization}. Finally, our survey adopts a more analytical perspective. We integrate the discussed CGL research within a unified mathematical framework, providing a detailed exposition of CGL settings and methodologies with an emphasis on technical precision. 
\color{black}

\section{Problem Setups}\label{sec:Problem Setups}

\begin{figure*}
    \centering
    \includegraphics[width=1.\textwidth]{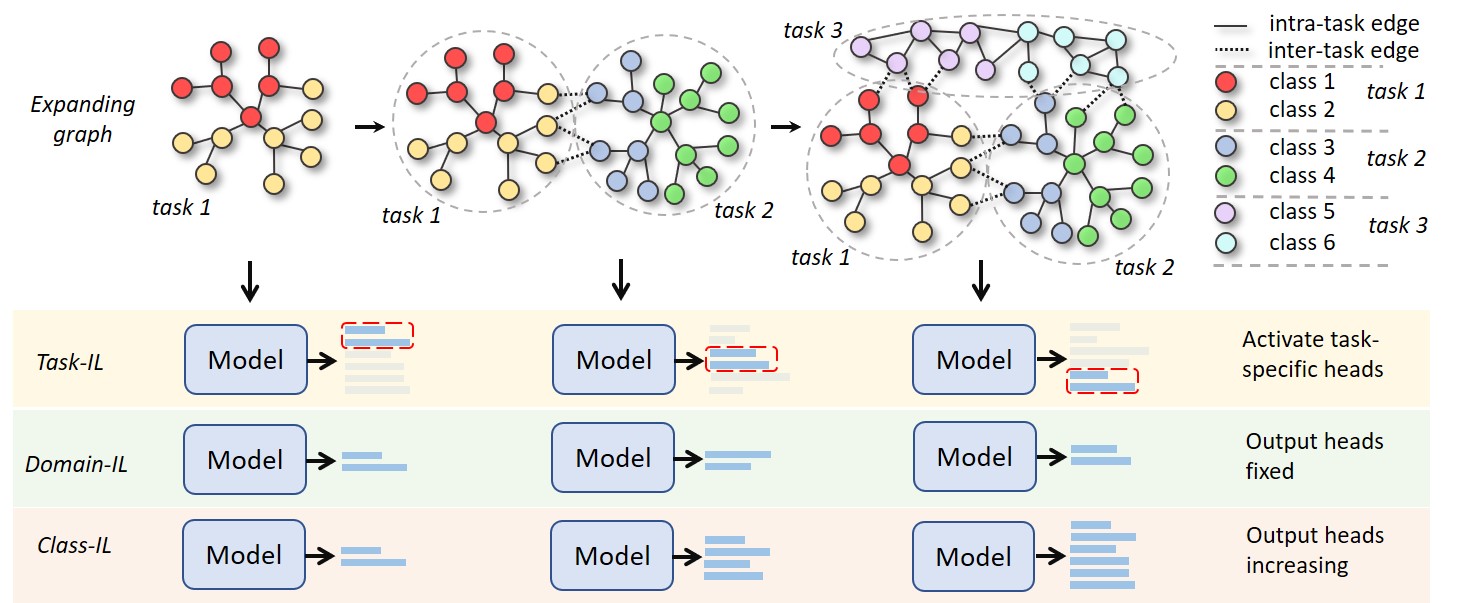}
    \caption{Illustration of the different incremental settings.}
    \label{fig:different_setting}
\end{figure*}

To cover different graph evolving scenarios, the continual learning process can be formulated on a sequence of graph tasks: 
$\mathcal{S}$ = $\{\mathcal{G}_1,\mathcal{G}_2, ..., \mathcal{G}_T\}$. Each graph task $\mathcal{G}_{\tau}$ represents the entire graph (or a set of independent graphs) that has grown (or increased) from task $1$ to $\tau$. Each $\mathcal{G}_{\tau}$ consists of a node set $\mathbb{V}_{\tau}$ containing all the nodes, and an edge set $\mathbb{E}_{\tau}$ containing all the edges. $\mathbb{E}_{\tau}$ can also be represented as an adjacency matrix $\mathbf{A}_{\tau}\in \mathbb{R}^{|\mathbb{V}_{\tau}|\times|\mathbb{V}_{\tau}|}$ (connected graphs for node-level CGL task and disconnected independent graphs for graph-level CGL tasks). Each entry $\mathbf{A}_{\tau}^{u,v}$ denotes the edge between node $u$ and $v$ ($\mathbf{A}_{\tau}^{u,v}=0$ if node $u$ and $v$ are not connected by an edge). The number of edges connected to a node is referred to as its degree, and the degree of all nodes with in a graph can be stored as the diagonal entries of a degree matrix $\mathbf{D}_{\tau}\in \mathbb{R}^{|\mathbb{V}_{\tau}|\times|\mathbb{V}_{\tau}|}$, \textit{i.e.} $\mathbf{D}_{\tau}^{u,u}$ is the degree of node $u$. In practice, $\mathbf{D}_{\tau}$ is often used to normalize the adjacency matrix, \textit{i.e.} $\hat{\mathbf{A}}_{\tau} = \mathbf{D}_{\tau}^{-\frac{1}{2}}\mathbf{A}_{\tau}\mathbf{D}_{\tau}^{-\frac{1}{2}}$. 

To conveniently refer to the nodes coming in different tasks, we denote the new nodes in a task $\tau$ as $\mathbb{V}_{\tau}^{new}$. Accordingly, we denote the induced subgraph based on $\mathbb{V}_{\tau}^{new}$, \textit{i.e.} all nodes in $\mathbb{V}_{\tau}^{new}$ together with all edges that connect nodes in $\mathbb{V}_{\tau}^{new}$, as $\mathcal{G}_{\tau}^{new}$. $\mathcal{G}_{\tau}^{new}$ is the task-specific subgraph of the new task $\tau$. Then, all the edges within each task specific subgraph are referred to as intra-task edges~\cite{zhang2022cglb}, while the other edges are inter-task edges connecting nodes across different tasks. We denote the intra-task edges of task $\tau$ as $\mathbb{E}_{\tau}^{intra}$, and the inter-task edges after the arrival of task $\tau$ as $\mathbb{E}_{\tau}^{inter} = \mathbb{E}_{\tau} \backslash \bigcup_{i=1}^\tau \mathbb{E}_{\tau}^{intra}$.

The formulation in this subsection applies to both node/edge-level CGL and graph-level CGL, which will be detailed in Section~\ref{sec:different granularity}. For node/edge-level CGL, which typically involves expanding networks (\textit{e.g.} citation networks), each graph $\mathcal{G}_{\tau}$ is the entire network that has grown from task $1$ to $\tau$. While for graph-level CGL, which typically involves independent graphs (\textit{e.g.} molecule graphs), each graph $\mathcal{G}_{\tau}$ is a set of disconnected independent graphs.

\begin{figure*}
    \centering
    \includegraphics[width=1.\textwidth]{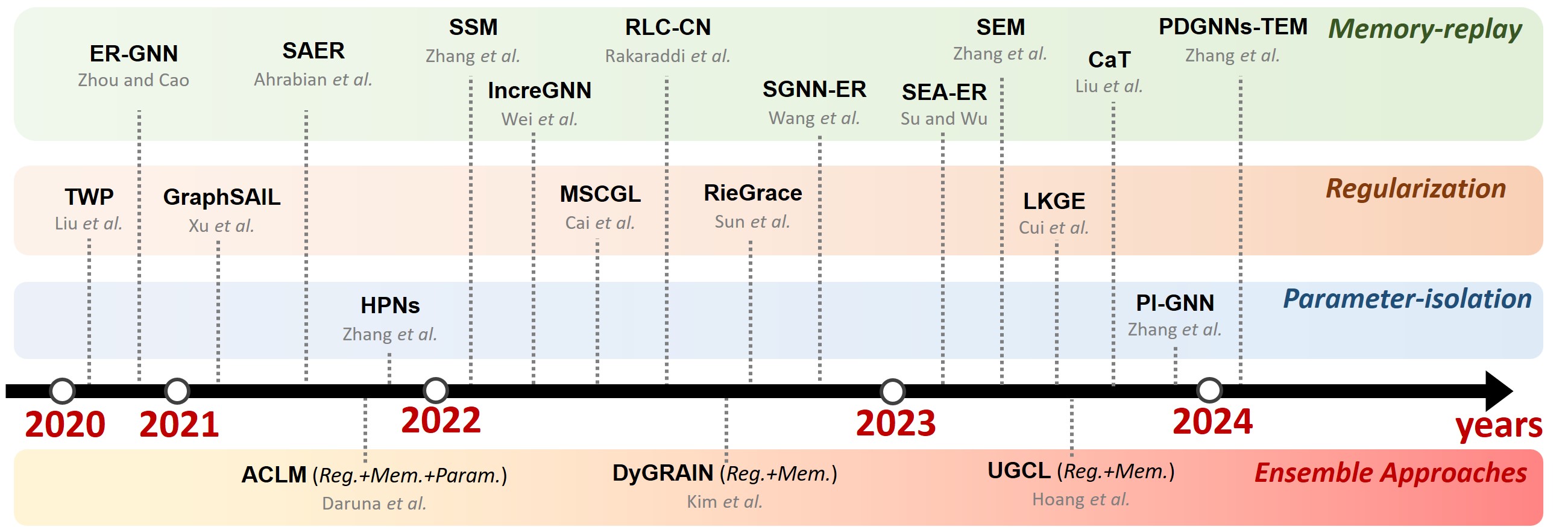}
    \caption{Timeline of the development of different CGL techniques.}
    \label{fig:timeline}
\end{figure*}

\subsection{Different Prediction Granularity}\label{sec:different granularity}
Different from learning on independent data, graph related tasks have different granularity that corresponds to different real-world scenarios and requires different task sequence constructions. 

Node/edge-level learning focuses on generating representations and predictions for individual nodes/edges. For example, given a citation network, classifying the papers is modeled as a node classification task (node-level). With a protein interaction network, in which the nodes are proteins and the edges are interactions between proteins, predicting the missing interactions is modeled as a link prediction (edge-level) task. Since nodes and edges belong to the same granularity and edge-related tasks are often based on the learnt node representation, we will use node-level to denote both node- and edge-level tasks in the following for simplicity. 

Formally, depending on whether inter-task edges are allowed to be preserved, node-level tasks can have two different formulations. First, without the inter-task edges, the representation of a task $\tau$ is:
\begin{align}
    \{\mathbf{h}_v | v\in \mathbb{V}_{\tau}^{new} \} = \mathrm{f}( \mathcal{G}_{\tau}^{new} ;\boldsymbol{\theta}_{\tau}),
\end{align}
where $\mathbf{h}_v$ is the representation of node $v$ generated by the model $\mathrm{f}( \cdot ;\boldsymbol{\theta}_{\tau})$ parameterized by $\boldsymbol{\theta}_{\tau}$ (the model parameters during learning task $\tau$). 
When the inter-task edges are preserved, then the input for learning task $\tau$ is not limited to $\mathcal{G}_{\tau}^{new}$. Since GNNs could aggregate information from previously observed nodes via the inter-tasks, the input becomes the entire graph $\mathcal{G}_{\tau}$, \textit{i.e.},
\begin{align}
    \{\mathbf{h}_v | v\in \mathbb{V}_{\tau}^{new} \} = \mathrm{f}( \mathcal{G}_{\tau} ;\boldsymbol{\theta}_{\tau}).
\end{align}

When dealing with node-level tasks, the incoming subgraph $\mathcal{G}_{\tau}^{new}$ is typically a connected graph. While for graph-level tasks, $\mathcal{G}_{\tau}^{new}$ is a set of disconnected graphs, \textit{e.g.}, a set of molecule graphs. In this case, the graph-level representations can be formulated as:
\begin{align}
    \{\mathbf{h}_g | g\in \mathcal{G}_{\tau}^{new} \} = \mathrm{f}( \mathcal{G}_{\tau}^{new} ;\boldsymbol{\theta}_{\tau}),
\end{align}
where $g$ denotes the disconnected components in $\mathcal{G}_{\tau}^{new}$.
Since graph-level CGL deals with individual graphs (disconnected components), there is no inter-task edge across different tasks. 

\subsection{Different Incremental Scenarios} 

According to whether the task identities are provided during testing and whether a model is required to figure out the task identities, CGL, similar to classic continual learning, can be categorized into task-incremental learning (task-IL), domain-incremental learning (domain-IL), and class-incremental learning (class-IL), as illustrated in Figure~\ref{fig:different_setting}.

\subsubsection{Task-IL}

In task-IL, the task identities are revealed to the model during testing, therefore the model is not required to identify the given tasks. For classification tasks, existing models typically increase their output dimensions to accommodate new tasks (\textit{e.g.}, new classes), and only activate the corresponding dimensions for each given task during testing. For example, in molecular property prediction tasks, each new task could be predicting whether a given molecule exhibits a certain new property. 

\subsubsection{Domain-IL}
Domain-IL refers to a scenario where the domain of the data changes across tasks, while the task remains static.
Therefore, the semantic meaning of a model's output dimensions is fixed. For example, one possible scenario for continual learning on knowledge graphs is to sequentially learn on graphs with different entities and relations, while the prediction task is always the completion of the triplets. Another example is the data split according to the temporal orders. The task could be the same across different temporal periods, while different temporal periods may contain data following significantly diverse regimes (domains).

\subsubsection{Class-IL}
Class-IL is the most challenging among the three scenarios. During testing, task identities are inaccessible and the model has to identify the given tasks. For classification tasks, a model typically increases the output dimensions to allocate new classes when they arrive and has to pick up the correct class among all learnt classes, unlike task-IL which only requires distinguishing between classes within a known task.

\section{Methods}\label{sec:Methods}

\begin{table*}[]
    \centering
    \caption{Summarization of different CGL techniques.}
    \begin{tabular}{c|c|c|c|p{6cm}}
    \toprule
    Method  & Applications & Task Granularity & Technique & Characteristics    \\ \midrule
    TWP \cite{liu2021overcoming}  & General & Node/Graph  & Reg. & Preserve the topology learnt from previous tasks \\
    RieGrace\cite{sun2023self}  & General & Node & Reg. & Maintain previous knowledge via knowledge distillation \\
    GraphSAIL~\cite{xu2020graphsail}  & Recommender Systems & Node & Reg. & Local and global structure preservation, node information preservation  \\
    MSCGL~\cite{cai2022multimodal}  & General & Node & Reg. & Parameter changes orthogonal to previous parameters \\
    LKGE~\cite{cui2023lifelong} &  Knowledge Graph & Node  & Reg. & Alleviating forgetting issue with l2 regularization\\ \midrule
    ER-GNN \cite{zhou2021overcoming}  & General & Node & Mem. & Replay representative nodes  \\
    SSM~\cite{zhang2022sparsified}  & General & Node & Mem. & Replay representative sparsified computation subgraphs \\
    SEM~\cite{zhang2023ricci} & General & Node & Mem. & Sparsify computation subgraphs based on information bottleneck \\
    PDGNNs-TEM~\cite{zhang2024topology}  &General & Node & Mem. & Replay representative topology-aware embeddings \\
    IncreGNN~\cite{wei2022incregnn}  & General & Node & Mem. & Replay nodes according to their influence\\
    RLC-CN~\cite{rakaraddi2022reinforced}  &General & Node & Mem. & Model structure adaption and dark experience replay \\
    SGNN-ER~\cite{wang2022streaming}  & General & Node  & Mem. & Model retraining with generated fake historical data \\
    SAER~\cite{ahrabian2021structure}  & Recommender System & Node & Mem. & Buffer the representative user-item pairs based on reservoir sampling\\
    SEA-ER~\cite{su2023towards_seaer} &  General & Node  & Mem. & Minimize the structural difference between the memory buffer and the original graph\\
    CaT~\cite{liu2023cat} &  General & Node  & Mem. & Train the model solely on balanced condensed graphs from all tasks \\ \midrule
    HPNs~\cite{9808404}  &  General & Node  & Para. & Extracting and storing basic features to encourage knowledge sharing across tasks, model expanding to accommodate new patterns \\
    PI-GNN~\cite{zhang2023continual}  &  General & Node  & Para. & Separate parameters for encoding stable and changed graph parts  \\ \midrule
    DyGRAIN~\cite{kim2022dygrain}  &  General & Node  & Mem.+Reg. & Alleviate catastrophic forgetting and concept shift of previous task nodes via memory replay and knowledge distillation \\
    ACLM~\cite{daruna2021continual} &  Knowledge Graph & Node  & Mem.+Reg.+Para. & Adapting general CL techniques to CGL tasks  \\
    UGCL~\cite{hoang2023universal} &  General & Node/Graph  & Mem.+Reg. & Memory replay \& local/global structure preservation\\             
    \bottomrule
    \end{tabular}
    \label{tab:summarization}
\end{table*}

Similar to traditional continual learning, CGL also approaches the problem from the perspectives of restricting the change in the model parameters, isolating and protecting the parameters that are important for previously learnt tasks, and replaying representative data from previous tasks to remind the model of the previously learnt patterns (Figure~\ref{fig:timeline}). However, a key challenge of CGL is the necessity to properly preserve the topological structure of the data, which is crucial information contained in the graph data. In this section, we will introduce the CGL techniques following this categorization, which is also summarized in Table~\ref{tab:summarization}.

\subsection{Regularization based methods}~\label{sec:regularization based methods}
Since the reason of the forgetting is that the model parameters trained for previous tasks are modified after being adapted to new tasks, traditional regularization based methods~\cite{kirkpatrick2017overcoming,aljundi2018memory} add penalty terms to prevent the parameters from being drastically changed. However, these methods do not explicitly preserve the topology of the graph data. Targeting this insufficiency, topology-aware weight preserving (TWP) \cite{liu2021overcoming} proposes to explicitly preserve the topology learnt on previous tasks via regularization on the model weights.

\begin{figure}
    \centering
    \includegraphics[width=0.4\textwidth]{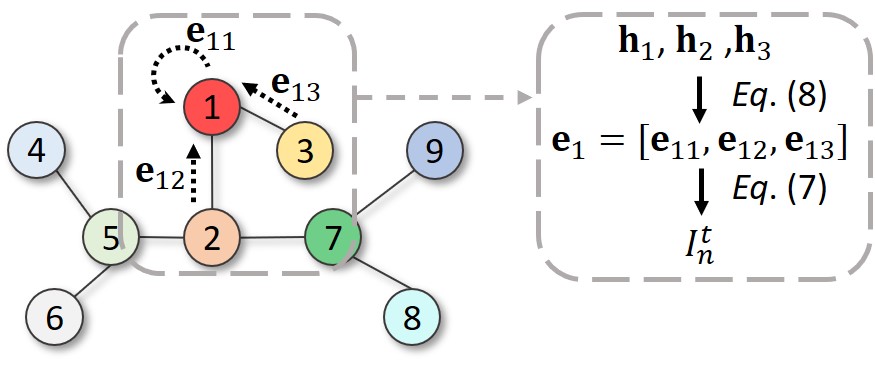}
    \caption{Topology preservation design in TWP~\cite{liu2021overcoming}.} 
    \label{fig:twp}
\end{figure}

Denoting the model parameters after learning the $t$-th task as $\boldsymbol{\theta}_t$, the learning of regularization based methods on the $(t+1)$-th can be generally formulated as,
\begin{align}\label{eq:regularization}
    \boldsymbol{\theta}_{t+1} = \argmin_{\boldsymbol{\theta}} \mathcal{L}_{t+1} + \sum_{n=1}^t I_n \otimes (\boldsymbol{\theta} - \boldsymbol{\theta}^*_n)^2.
\end{align}
In Equation~\ref{eq:regularization}, $\boldsymbol{\theta}^*_n$ is the optimal parameters for the $n$-th task, and the importance scores $I_n$ indicate the importance of each parameter for the performance in task $n$. In other words, the change of the parameters that are important to previous tasks are strongly penalized, while the less important parameters are more free to adapt to the new tasks. In different methods, $I_n$ are calculated with different strategies. In TWP, the importance scores consist of two parts. 
The first part is calculated for each weight of the model based on the sensitivity of the loss on the weight, which is measured with the gradient of the loss with respect to the weight. This part resembles the strategies adopted by EWC~\cite{kirkpatrick2017overcoming} and MAS~\cite{aljundi2018memory}, and serves to preserve the weights that are important to previous tasks. We denote the corresponding importance scores as $I^p_n$.
The second part, which is the key contribution of TWP, is the sensitivity of the learnt graph structure on the weight. For backbones like graph attention network (GAT) that explicitly learns the strength of the edges via the attention scores, the learnt structures are deemed as the attention score on each weight, and the sensitivity is measured as the gradient of the squared \textit{l}$_2$ norm of the attention scores with respect to the weights. Denoting the attention score between two nodes $i$ and $j$ as $e_{ij}$, the attention score between a node $i$ and its all neighbors can be represented as $\mathbf{e}_i = [e_{i1},...,e_{i|\mathcal{N}_i|}]$. After that, the sensitivity (importance) of the topology learnt in task $n$ with respect to the model parameters is defined as the gradient of the squared $l2$ norm with respect to the parameters (Figure~\ref{fig:twp}).
\begin{align}
    I_n^{t} = \frac{\partial \big( \big| [\mathbf{e}_1, ..., \mathbf{e}_{|V^n|}] \big| \big)}{\partial \boldsymbol{\theta}}
\end{align}
For backbones without attention mechanism, the attention score between two nodes is calculated with a non-parametric attention mechanism,
\begin{align}
    e_{ij} = \mathbf{h}_i^T\tanh(\mathbf{h}_j)
\end{align}
With the importance scores for both the model performance and the topology, the final importance is obtained by their weighted summation,
\begin{align}
    I_n = \lambda_p I_n^p + \lambda_t I_n^t.
\end{align}

Besides the learnt attention scores, the graph curvature is another important topological property of the graph data, and is crucial for graph learning~~\cite{topping2021understanding,ye2019curvature}. When learning on a sequence of graphs, considering that the curvature of the incoming graph may constantly change, Riemannian Graph Continual Learner (RieGrace)\cite{sun2023self} further proposes Adaptive Riemannian GCN (AdaRGCN) for accommodating the change in the graph curvature and Label-free Lorentz Distillation for alleviating the forgetting problem. Specifically, AdaRGCN can constantly adapt and capture the topology of the given graphs with different curvatures. And the Label-free Lorentz Distillation serves to maintaining the learnt knowledge when learning on the graph sequence. Considering the possible scarcity of labels in real worlds applications, Label-free Lorentz Distillation trains a model based on contrastive learning without any label. It consists of two parts, including a cross-layer intra-distillation and a cross-model inter-distillation. The intra-distillation serves to learn the pattern of the current graph, and the contrastive learning is conducted by maximizing the agreement between the representations of each node at different layers of the model. Formally, the objective function is
\begin{align}\label{eq:intra-dis}
    \mathcal{J}(\mathbf{x}_i^{s,L},\mathbf{x}_i^{s,H})=\log\frac{\exp \mathrm{Sim}^{\mathcal{L}}(\mathbf{x}_i^{s,L},\mathbf{x}_i^{s,H})}{\sum_{j=1}^{|\mathcal{V}|}\mathbb{I}\{i\neq j\}\exp \mathrm{Sim}(\mathbf{x}_i^{s,L},\mathbf{x}_i^{s,H})}.
\end{align}
In Equation~\ref{eq:intra-dis}, $\mathbf{x}_i^{s,L}$ and $\mathbf{x}_i^{s,H}$ denote the low- and high-level representations of the $i$-th node, extracted by shallow and deep layers of the model. Since these two representations correspond to the same node, the objective is designed to maximize their similarity, which is measured by the adopted function $\mathrm{Sim}^{\mathcal{L}}(\cdot,\cdot)$. And $\mathcal{V}$ is the node set of the entire graph. 
Besides learning the pattern of the current graph, the knowledge learnt from the previous tasks should also be distilled into the current model, which is achieved by the inter-distillation loss,
\begin{align}\label{eq:inter-dis}
    \mathcal{J}(\mathbf{x}_i^{t,H},\mathbf{x}_i^{s,H})=\log\frac{\exp \mathrm{Sim}^{\mathcal{L}}(\mathbf{x}_i^{t,H},\mathbf{x}_i^{s,H})}{\sum_{j=1}^{|\mathcal{V}|}\mathbb{I}\{i\neq j\}\exp \mathrm{Sim}(\mathbf{x}_i^{t,H},\mathbf{x}_i^{s,H})}.
\end{align}
Equation~\ref{eq:inter-dis} has the same formulation as Equation~\ref{eq:intra-dis}, while the difference is that the contrastive learning is between the high-level representation of the teacher model and the student model. These two loss on each node are then summed up and balanced with a parameter $\lambda$, $\mathcal{J}_{overall}=\sum_{i=1}^{|\mathcal{V}|}\mathcal{J}(\mathbf{x}_i^{s,L},\mathbf{x}_i^{s,H}) + \lambda \sum_{i=1}^{|\mathcal{V}|}\mathcal{J}(\mathbf{x}_i^{t,H},\mathbf{x}_i^{s,H})$. By optimizing these two losses simultaneously, the current student model contains knowledge from both the current graph and the previously learnt model, and will serve as the teacher model when learning the next task.

Besides RieGrace, distillation is also adopted by Universal Graph Continual Learning (UGCL)~\cite{hoang2023universal}. UGCL is a technique based on both memory replay and knowledge distillation. The memory replay module is constructed by storing representative nodes as well as the associated neighborhood in a buffer, which are then replayed to the model when learning on new tasks to avoid forgetting. After learning each task, the memory buffer will be updated with two steps. First, all of the observed data of the current task will be added into the buffer. If the resulting buffer size exceeds the budget, a replacement process will be triggered. The replacement process will iteratively remove one random example from the largest class in the buffer until the buffer size is not larger than the budget.
The knowledge distillation module of UGCL is designed to preserve the learned topological information from both local and global perspectives. For local structure, UGCL adopts the strategy to preserve the difference between a node and its neighborhood. Denoting the representation of a node $v$ generated by the current/old models as $\mathbf{z}_v^c$/$\mathbf{z}_v^o$, the local structure between a node $v$ and its neighbourhood $\mathcal{N}(v)$ in a buffered graph $g$ is formulated as,
\begin{align}
    S^c_{(g,v)} = \mathbf{z}_v^c - \frac{1}{|\mathcal{N}(v)|} \sum_{w \in \mathcal{N}(v)} \mathbf{z}_w^c \\
    S^o_{(g,v)} = \mathbf{z}_v^o - \frac{1}{|\mathcal{N}(v)|} \sum_{w \in \mathcal{N}(v)} \mathbf{z}_w^o.
\end{align}
Based on the local structure difference scores of both the current and old models ($S^c_{(g,v)}$, $S^c_{(g,v)}$), the objective for preserving the local structure difference is,
\begin{align}
    \mathcal{L}^{LS} = \frac{1}{K_g \cdot K_n} \sum_{j=1}^{K_g} \sum_{i=1}^{K_v} [1-\mathrm{cos}(S^c_{(j,i)},S^o_{(j,i)})],
\end{align}
where $K_g$ and $K_v$ denote the number of buffered graphs and the number of nodes selected from the buffered graphs.
Finally, to preserve the global structure, the objective is formulated as minimizing the cosine distance between the graph-level representations output from the current and old models,
\begin{align}
    \mathcal{L}^{GS} = \frac{1}{K_g} \sum_{j=1}^{K_g} [1-\mathrm{cos}(\mathbf{g}_j^o,\mathbf{g}_j^c)],
\end{align}
where $\mathbf{g}_j^o$/$\mathbf{g}_j^c$ denotes the representations of graph $j$ generated by the current and old model.

The methods introduced above mainly focus on the forgetting problem. However, different from traditional continual learning, in which the new and old data are independent, the newly incoming data (graph nodes) in CGL could form edges connecting to the existing nodes. Since GNNs generate the prediction for a node based on its multi-hop neighbors, the newly formed connections between new and old nodes will cause concept drift for the old nodes and alter their representation~\cite{zhang2022cglb}.  
Therefore, targeting both this challenge and the catastrophic forgetting problem, DyGRAIN~\cite{kim2022dygrain} is designed to first identify the nodes that are most influenced by new nodes and the nodes that are most vulnerable to forgetting. After that, model retraining and knowledge distillation are applied to alleviate the influence of the new nodes and the forgetting problem. For the detection of the vulnerable nodes, the changing neighborhood problem is termed as the time-varying receptive field in the DyGRAIN paper. Based on the neighborhood aggregation of GNNs, the nodes most influenced by the changed receptive field are detected based on the adjacency matrix to conduct the neighborhood aggregation. Specifically, denoting the previously observed nodes as $\mathcal{M}$, the nodes in the $t$-th task as $\mathcal{V}^{(t)}$, an indicator matrix $\mathbf{V}$ can be derived to represent the influence of each node after certain rounds of message passing.
\begin{equation}\mathbf{V}_{ii}^{(0)}=
\begin{cases}
1 \quad if \quad i \in \mathcal{V}^{(t)}\\
0 \quad if \quad i \in \mathcal{M},
\end{cases}
\mathbf{V}^{(l)} = \mathbf{P}^{(t)}\mathbf{V}^{(l-1)},
\end{equation}
where $\mathbf{P}^{(t)} = \mathbf{A}^{(t)}\mathbf{D}^{-1}$ is the adjacency matrix $\mathbf{A}^{(t)}$ normalized by the degree matrix $\mathbf{D}$. By iteratively applying $\mathbf{P}^{(t)}$ for $L$ times, the entries in $\mathbf{V}^{(L)}$ denote how much information (\textit{i.e.} influence) would be propagated from each node to the $L$-hop neighbors. Accordingly, a Structural Influence (SI) score of a node $i$ is defined as:
\begin{align}
    \pi_{SI}(i)  = \sum_{j \in \mathcal{M}\cup \mathcal{V}^{(t)}} \sum_{l=1}^L \mathbf{V}_{ij}^{(l)}.
\end{align}
Based on the structural influence score defined above, a subset of $k$ nodes with the highest scores (\textit{i.e.} $\mathcal{S}_{struct}$) are selected as the ones that are most vulnerable to the influence of the new nodes. Besides the purely structural influence, the authors also consider the influence based on the change in the node embeddings. Denoting the GNN model as $\mathrm{f}_{\boldsymbol{\theta}}(\cdot)$, the representation of a node $i$ after the arrival of task $t$ can be denoted as, $\mathbf{h}_i^{(t)} = \mathrm{f}_{\boldsymbol{\theta}}(\mathbf{x}_i, \mathbf{A}^{(t)};\boldsymbol{\theta}^{(t-1)})$. And the influence of the $t$-th task is derived based on the change in the node embeddings and defined as the Feature Influence (FI) score,
\begin{align}
    \pi_{FI}(v_i) = 1-\frac{\mathbf{h}_i^{(t-1)} \cdot \mathbf{h}_i^{(t)}}{||\mathbf{h}_i^{(t-1)}|| \cdot ||\mathbf{h}_i^{(t)}||}.
\end{align}
Similarly, with the obtained feature influence score, the top $k$ nodes, \textit{i.e.} $\mathcal{S}_{feat}$, are also selected as the vulnerable ones. Above all, to mitigate the influence from the time-varying receptive field, the union of the two sets of nodes, \textit{i.e.} $\mathcal{S}_{IP} = \mathcal{S}_{struct} \cup \mathcal{S}_{feat}$, will be included in the memory for retraining the model.

Similar to the procedure to mitigate the influence of time-varying receptive field, the catastrophic forgetting is also dealt with by first identifying the most vulnerable nodes. Since the nodes within the $L$-hop neighbors are already sampled, they will be excluded during the identification of the nodes that are vulnerable to catastrophic forgetting, and the remaining node is denoted as $\mathcal{V}_{C}$. With $\mathcal{V}_{C}$, the loss on the previous task is,
\begin{align}
    \mathcal{L}^{(t-1)} = l(\mathcal{V}_C, \tilde{A}^{(t-1)}; \mathrm{f}_{\boldsymbol{\theta}^{(t-1)}}).
\end{align}
Since lower loss values correspond to the more representative nodes (the model is well adapted for them), from $\mathcal{V}_{C}$, a subset of nodes is selected by lower $q$ percentile according to the loss value, $\mathcal{V}' = \{ i \in \mathcal{V}_C | \mathcal{L}_i^{(t-1)} \leq P_q(\mathcal{L}^{(t-1)}) \}$. To find the most vulnerable nodes, the first step is to estimate the new loss in task $t$ of the nodes $\mathcal{V}'$ as,
\begin{align}
    \Hat{\mathcal{L}^{(t)}} \approx l(\mathcal{V}',\tilde{\mathbf{A}}^t;\mathrm{f}_{\boldsymbol{\theta}^{(t-1)}}).
\end{align}
Then the loss-based importance score of a node $i$ is defined as the increase in the loss,
\begin{align}
    \pi_{CF}(i) = \Hat{\mathcal{L}}_i^{(t)} - \mathcal{L}^{(t-1)}_i,
\end{align}
and the top $k$ nodes are the ones that with the largest loss and are selected as the most vulnerable ones to the new nodes, \textit{i.e.} $\mathcal{S}_{KD}$. The role of $\mathcal{S}_{KD}$ is two fold. First, knowledge distillation is conducted to protect the performance of these vulnerable nodes by minimizing the discrepancy between the node representations generated by the new and old models. Second, $\mathcal{S}_{KD}$ is also used for retraining the model together with the nodes in $\mathcal{S}_{IP}$. Above all, by retraining the model with the nodes in $\mathcal{S}_{IP}$, DyGRAIN is capable of ensuring that the old nodes whose neighborhood (receptive field) is altered by new nodes with up-to-date representations. Through the knowledge distillation and model retraining with nodes in $\mathcal{S}_{KD}$, the catastrophic forgetting can also be alleviated.

Unlike the grid data, \textit{e.g.} images, videos, and text, graph data has tremendous different branches with significantly different properties. Accordingly, besides the above works for general CGL on any graph data, there are also methods specialized for a certain kind of graphs. For example, specially designed for incremental learning for recommender systems, Graph Structure Aware Incremental Learning (GraphSAIL) designs three knowledge distillation techniques to avoid the forgetting problem, including local structure distillation, global structure distillation, and self-embedding distillation. Since the nuclear operation of GCNs is the neighborhood aggregation over the local neighbors, preserving the local contextual structure is crucial. In GraphSAIL, this local structure is reflected through the affinity scores between the center nodes and their neighbors. And the affinity is maintained by knowledge distillation between the new and old tasks. Mathematically, it can be formulated as,
\begin{align}
    \nonumber \mathcal{L}_{local} = \Big( \frac{1}{|\mathcal{U}|\sum_{u\in \mathcal{U}}(\mathbf{e}_u^{(t-1)}\cdot \mathbf{c}_{u,N_u^{t-1}} - \mathbf{e}_u^t \cdot \mathbf{c}_{u,N_u^{t-1}}^t )^2} \\ + \frac{1}{|I|} \sum_{i \in \mathcal{I}} (\mathbf{e}_i^{t-1} \cdot \mathbf{c}_{i,N_i^{t-1}}^t)^2 \Big),
\end{align}
where $\mathcal{U}$ and $\mathcal{I}$ are the user set and item set, $\mathbf{e}_u^t$ is the embedding of the user $u$ during learning task $t$, and $c$ denotes the average of the neighboring user nodes or item nodes,
\begin{align}
    \mathbf{c}_{u,N_u^{t-1}} = \frac{1}{|\mathcal{N}_u^{(t-1)}|} \sum_{i' \in N_u^{t-1}} \mathbf{e}_{i'}^t, \\
    \mathbf{c}_{i,N_u^{t-1}} = \frac{1}{|\mathcal{N}_i^{(t-1)}|} \sum_{u' \in N_i^{t-1}} \mathbf{e}_{u'}^t.
\end{align}
Besides the local structure denoting the context of a node, the global position of a node within the graph is also crucial. For example, the distance among the users can reflect certain preference groups. Therefore, a global structure distillation is also designed. Specifically, to capture the global structure, a set of anchor nodes is first calculated for both the users and the items based on K-means clustering algorithm. Then each node has a similarity distribution over all the anchors denoting the relative position of the node within the graph,
\begin{align}
    GS^t_{u,\mathcal{A}_u^{t,k}} = \frac{e^{\mathrm{SIM}(\mathbf{e}_u^t, \mathcal{A}_u^{t,k})}/ \tau}{\sum_{k'=1}^K e^{\mathrm{SIM}(\mathbf{e}_u^t, \mathcal{A}_u^{t,k'})/ \tau}}, \mathrm{SIM}(\mathbf{a},\mathbf{b}) = \mathbf{a}^\mathrm{T}\mathbf{b},
\end{align}
where $\mathcal{A}_u^{t}$ denotes the anchors calculated during task $t$. The goal of global structure preserving is to maintain the similarity distribution across different tasks,
\begin{align}
    \mathcal{S}_{u, \mathcal{A}_u} = D_{KL}(GS^t_{u,\mathcal{A}_u^{t}} || GS^{t-1}_{u,\mathcal{A}_u^{t-1}}).
\end{align}
Finally, besides the topological information including both the local and global structures, the information contained in each user/item node is also highly valuable. To protect this part of information, a self-distillation is proposed,
\begin{align}
   \nonumber  \mathcal{L}_{self} = \big( \frac{1}{|\mathcal{U}|} \sum_{u \in \mathcal{U}} \frac{\eta_u}{||\eta_U||_2} ||\mathbf{e}_u^{t-1} - \mathbf{e}_u^t||_2 \\ +\frac{1}{|\mathcal{I}|} \sum_{i \in \mathcal{I}} \frac{\eta_i}{||\eta_I||_2} ||\mathbf{e}_i^{t-1} - \mathbf{e}_u^t||_2 \big),
\end{align}
where $\eta_u = \frac{|\mathcal{N}_u^{t-1}|}{|\mathcal{N}_u^{t}|}$ and $\eta_i = \frac{|\mathcal{N}_i^{t-1}|}{|\mathcal{N}_i^{t}|}$ are the normalizing factors. 

Multi-modal Structure-evolving Continual Graph Learning (MSCGL)~\cite{cai2022multimodal} is proposed to tackle the challenges of CGL with multi-modal data. Specifically, MSCGL consists of two parts, including a Neural Architecture Search (NAS) module serving to adaptively optimize the model architecture for accommodating new tasks, and a Group Sparse Regularization (GRS) module for preserving crucial information of the previously learnt tasks. The search space of the model architecture is designed as multiple choices of different GNNs, in which the different modalities are processed with separate GNNs. Denoting the distribution of the model architecture $\mathrm{a}$ as $P(\mathrm{a};\boldsymbol{\theta})$ parameterized by $\boldsymbol{\theta}$, the objective can be formulated as a bi-level optimization including the maximization of the expected accuracy $\mathcal{E}[\mathcal{R}(\mathrm{a}(w*,G))]$ and minimization of the training loss $\mathcal{L}_{train}(\mathrm{a}(w,G))$,
\begin{align}
    &\max \mathcal{E}[\mathcal{R}(a(w*,G))],& \\
    &s.t. \quad  w* = \argmin_w \mathcal{L}_{train}(\mathrm{a}(w,G)).& \nonumber
\end{align}
With the model obtained from NAS, a Group Sparse Regularization (GSR) is proposed to sparsify the network. During the sparsification, two restrictions proposed by \cite{pasunuru2019continual} are adopted to ensure that the parameter changes are block-sparse and orthogonal to the previous parameters, so that the obtained new structure has less severe forgetting problem on the previous tasks.

In the area of knowledge graph research, Lifelong Knowledge Graph Embedding (LKGE)~\cite{cui2023lifelong} is constructed as a specialized technique for knowledge graph based on embedding transfer and regularization. Similar to the general CGL, the sequential learning process can still be formulated as training the model on a growing graph sequence $\mathcal{S}$ = $\{\mathcal{G}_1,\mathcal{G}_2, ..., \mathcal{G}_T\}$. Within the context of the knowledge graph, the graph is composed of facts, each of which is a tuple $(s,r,o)$ denoting the existence of a relation $r$ between the subject entity $s$ and object entity $o$. Besides, the new entities and relations emerging at the task $\tau$ are denoted as $\mathcal{E}_{\tau}$ and $\mathcal{R}_{\tau}$. LKGE mainly consists of three parts, including a masked knowledge graph autoencoder as the backbone model for knowledge graph embedding, an embedding transfer technique for transferring the learnt knowledge into unseen new entities and relations, and an embedding regularization module for alleviating the catastrophic forgetting problem. First of all, the masked knowledge graph autoencoder aims to capture information from both new and old tasks. Therefore, it is designed to minimize the distance between the embeddings and representations reconstructed by a masked neighborhood subgraph containing both old and potential new entities. 
\begin{align}\label{eq:LKGE_MAE}
 \mathcal{L}_{MAE} & = \sum_{e \in \mathcal{E}_{\tau}} ||\mathbf{e}_{\tau}-\bar{\mathbf{e}}_{\tau}|| + \sum_{r \in \mathcal{R}_{\tau}} ||\mathbf{r}_{\tau}-\bar{\mathbf{r}}_{\tau}||.
\end{align}
In Equation~\ref{eq:LKGE_MAE}, $\mathbf{e}_{\tau}$ and  $\mathbf{r}_{\tau}$ are the embeddings of an entity $e$ and a relation $r$ in task $\tau$. The masked autoencoder $\mathrm{MAE}(\cdot)$ takes all facts involving an entity $e$ / relation $r$ from the observed tasks, and then generate a representation $\bar{\mathbf{e}}_{\tau}=\mathrm{MAE}\big(\cup_{j=1}^{\tau} \mathcal{N}_j(e)\big)$ / $\bar{\mathbf{r}}_{\tau}=\mathrm{MAE}\big(\cup_{j=1}^{\tau} \mathcal{N}_j(r)\big)$ to encourage the embeddings to contain information from both new and old facts.
Then, inspired by the idea of TransE~\cite{bordes2013translating} ($\mathbf{s}+\mathbf{r}\approx\mathbf{o}$), the encoder is designed to represent the subject entity as the subtraction between object and relation ($\mathrm{f}_s(\mathbf{r},\mathbf{o}) = \mathbf{o}-\mathbf{r}$), the relation entity as the subtraction between object and subject ($\mathrm{f}_s(\mathbf{s},\mathbf{o}) = \mathbf{o}-\mathbf{s}$).
Accordingly, the encoders can be derived as:
\begin{align}
    \bar{\mathbf{e}}_{\tau} = \frac{\sum_{j=1}^{\tau}\sum_{(s,r,o)\in\mathcal{N}_j(e)}\mathrm{f}_s(\mathbf{r}_{\tau},\mathbf{o}_{\tau})}{\sum_{j=1}^{\tau}|\mathcal{N}_j(e)|} \\
    \bar{\mathbf{r}}_{\tau} = \frac{\sum_{j=1}^{\tau}\sum_{(s,r,o)\in\mathcal{N}_j(r)}\mathrm{f}_r(\mathbf{s}_{\tau},\mathbf{o}_{\tau})}{\sum_{j=1}^{\tau}|\mathcal{N}_j(r)|}.
\end{align}
However, in continual learning, the access to all previous data is now allowed, and the aim is approximated as,
\begin{align}
    \bar{\mathbf{e}}_{\tau} \approx \frac{\sum_{j=1}^{\tau-1}|\mathcal{N}_j(e)|\mathbf{e}_{\tau-1} +\sum_{(s,r,o)\in\mathcal{N}_j(e)}\mathrm{f}_s(\mathbf{r}_{\tau},\mathbf{o}_{\tau})}{\sum_{j=1}^{\tau}|\mathcal{N}_j(e)|} \\
    \bar{\mathbf{r}}_{\tau} \approx \frac{\sum_{j=1}^{\tau-1}|\mathcal{N}_j(r)|\mathbf{r}_{\tau-1} + \sum_{(s,r,o)\in\mathcal{N}_j(r)}\mathrm{f}_r(\mathbf{s}_{\tau},\mathbf{o}_{\tau})}{\sum_{j=1}^{\tau}|\mathcal{N}_j(r)|}.
\end{align}
The learning on the new task $\tau$ is also based on TransE, and is formulated as,
\begin{align}
    \mathcal{L}_{new} = \sum_{(s,r,o)\in D_{\tau}} \max(0,\gamma+\mathrm{f}(\mathbf{s},\mathbf{r},\mathbf{o})-\mathrm{f}(\mathbf{s}',\mathbf{r},\mathbf{o}')),
\end{align}
where $D_{\tau}$ contains all facts in task $\tau$, $\gamma$ is the margin hyperparameter, and $(\mathbf{s}',\mathbf{r}',\mathbf{o}')$ denotes the embedding of a negative fact, which is generated by replacing the entities in a real fact with random entities.
Second, the embedding transfer technique transfers the learned knowledge to unseen new entities by initializing the unseen entities as a aggefation of the learned entities,
\begin{align}
    \mathbf{e}_{\tau} = \frac{1}{|\mathcal{N}_{\tau}(e)|}\sum_{(e,r,o)\in\mathcal{N}_{\tau}(e)}\mathrm{f}_s(\mathbf{r}_{\tau-1},\mathbf{o}_{\tau-1}).
\end{align}
Third, a regularization term is designed for both entity and relation embeddings for alleviating the forgetting issue,
\begin{align}
    \mathcal{L}_{old}=\sum_{e\in \mathcal{E}_{\tau-1}}\omega(e)||\mathbf{e}_{\tau}-\mathbf{e}_{\tau-1}||_2^2+\sum_{r\in \mathcal{R}_{\tau-1}} \omega(r)||\mathbf{r}_{\tau}-\mathbf{r}_{\tau-1}||_2^2,
\end{align}
in which the contribution of the regularization terms are controlled by $\omega(\cdot)$. $\omega(\cdot)$ is calculated based on the number of observed data, \textit{i.e.} $\omega(e) = 1-\frac{|\mathcal{N}_{\tau}(e)|}{\sum_{j=1}^{\tau}|\mathcal{N}_j(e)|}$, $\omega(r) = 1-\frac{|\mathcal{N}_{\tau}(r)|}{\sum_{j=1}^{\tau}|\mathcal{N}_j(r)|}$.
Finally, the total loss function consists of all of the terms introduced so far,
\begin{align}
    \mathcal{L} = \mathcal{L}_{new} + \alpha\mathcal{L}_{old} + \beta\mathcal{L}_{MAE}
\end{align}

Besides the works on developing novel CGL techniques, researchers are also interested in exploring the effectiveness of existing continual learning techniques for specific CGL problems.
For example, in fake news detection tasks, to continually accommodating new data without retraining the model on all previous data , which is prohibitively expensive, \cite{han2020graph} adopts the Elastic Weight Consolidation (EWC)~\cite{kirkpatrick2017overcoming} and Gradient Episodic Memory (GEM)~\cite{lopez2017gradient}, which successfully protect the performance on both the previous data and new data empirically. Both EWC and GEM are model-agnostic continual learning methods, and do not explicitly consider the topological information within the graph data.

\subsection{Memory Replay based Methods}
\begin{figure}
    \centering
    \includegraphics[width=0.5\textwidth]{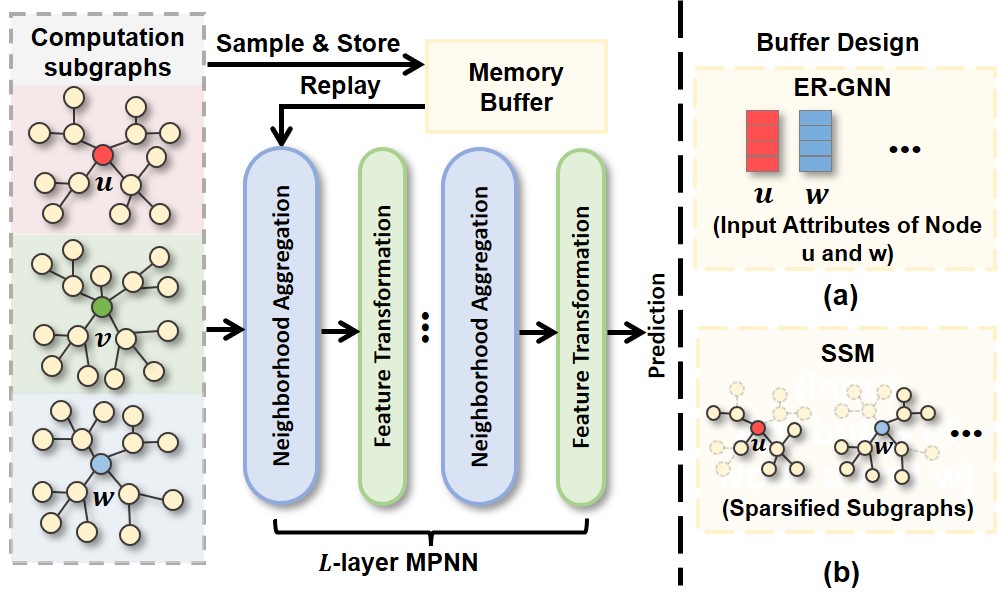}
    \caption{Pipelines of two representative memory replay based techniques. (a) ER-GNN~\cite{zhou2021overcoming} stores single nodes. (b) SSM~\cite{zhang2022sparsified} stores sparsified computation subgraphs.} 
    \label{fig:ergnn_ssm}
\end{figure}

Memory replay based methods prevent forgetting by retraining the model with representative data from previous tasks. Traditional continual learning methods process individual data without interactions, which can be simply sampled and stored in a buffer. However, for learning on graphs, memory replay based methods would meet the memory explosion challenge~\cite{zhang2022cglb}. When generating the representation of a node (a datum), Graph Neural Networks (GNNs) typically aggregate information from multi-hop neighbors. Therefore, to regenerate the representation of a single node, the information from an exponentially expanding neighborhood have to be stored. On dense graphs, the memory consumption would easily become intractable~\cite{zhang2022cglb}.
Due to this challenge, Experience Replay Graph Neural Network (ER-GNN) \cite{zhou2021overcoming} (Figure~\ref{fig:ergnn_ssm} (a)) chooses to ignore the graph topology and only store the attributes of one single node for regenerating the representation. Denoting the node buffer as $\mathcal{B}$, the loss function of ER-GNN when learning the $n$-th task can be formulated as
\begin{align}
    \mathcal{L}_n(\boldsymbol{\theta}, \mathcal{D}^{tr}_i, \mathcal{B}) = \beta \mathcal{L}_n(\boldsymbol{\theta}, \mathcal{D}^{tr}_i) + (1-\beta) \mathcal{L}_n(\boldsymbol{\theta}, \mathcal{B}),
\end{align}
where
\begin{align}
    \beta = \frac{|\mathcal{B}|}{|\mathcal{D}^{tr}_i| + |\mathcal{B}|}.
\end{align}
In other words, with the memory buffer, the total training loss is a weighted summation of the loss calculated with the data from the current task and the buffered data. And the contribution of these two parts are balanced based on their relative sizes. 
To properly populate the memory buffer, three different strategies are proposed in ER-GNN to select the representative nodes, including Mean of Feature (MF), Coverage Maximization (CM), and Influence Maximization (IM). The MF strategy is developed based on the intuition that the average feature vector of each class is representative, therefore the selected nodes should be the ones that are the closest to the average feature vector. The average feature vector (prototype) for a class $l$ can be obtained in two ways,
\begin{align}
    \mathbf{c}_l = \frac{1}{|V_l|}\sum_{(v \in V)}\mathbf{x}_v,
    \mathbf{c}_l = \frac{1}{|V_l|}\sum_{(v \in V)}\mathbf{h}_v,
\end{align}
where $V_l$ denotes the node set of the class $l$, $\mathbf{x}_v$ is the input attributes of the node $v$, and $\mathbf{h}_v$ is the representation of $v$ generated by the GNN.
The second strategy, CM, aims to maximize the diversity of the selected nodes. Specifically, for each node in each task, the number of nodes from different classes within a certain distance $d$ is first counted as,
\begin{align}\label{eq:CM}
    \mathcal{N}(v) = \{u|\mathrm{dist}(v,u)<d, \mathbf{y}_v \neq \mathbf{y}_u\}.
\end{align}
In equation~\ref{eq:CM}, $y_v$ denotes the label of node $v$, $\mathrm{dist}(\cdot,\cdot)$ measures the distance between two nodes, and $d$ is a specified threshold. The lower $|\mathcal{N}(v)|$ a node $v$ has, the more distant node $v$ is from the nodes from different classes. Therefore, to maximize the coverage of the selected nodes in the representation space, CM is designed to select the several nodes with the lowest $|\mathcal{N}(v)|$. 
The third strategy, IF, aims to find the nodes with the most influence on the model parameters. Accordingly, a naive approach to probe the influence of a node is removing it and retraining the model to check the parameter change, which is very inefficient. Therefore, based on an alternative approach~\cite{hampel1986robust} to replace the node removal with loss upweighting, the authors propose to approximate the importance of a training node by its influence on the loss of testing nodes. For a training node $v$ and a testing node $u$, the influence of node $v$ during task $n$ is derived as,
\begin{align}
    I(v,u) = - \nabla_{\boldsymbol{\theta}}\mathcal{L}_n(u)^T\mathbf{H}_{\boldsymbol{\theta}}^{-1}\nabla_{\boldsymbol{\theta}}\mathcal{L}_n(v),
\end{align}
where $\mathbf{H}_{\boldsymbol{\theta}}$ is the Hessian matrix of the loss function. Finally, the influence on all testing nodes are summed up as the approximation of the node $v$.

Although being easy to implement, the valuable topological information is not preserved in ER-GNN. To preserve the graph topology and maintain a tractable memory consumption at the same time, Sparsified Subgraph Memory (SSM) \cite{zhang2022sparsified} is proposed to store sparsified computation subgraphs. Given a computation subgraph to store, SSM would sample a fixed number of nodes iteratively from the 1-hop to a specified $K$-hop neighbors. This hop-by-hop sampling manner can ensure the connectivity of the sparsified subgraph. After storing the sparsified subgraphs in the memory buffer $\mathcal{SSM}$, the total loss for learning on a task $\tau$ is derived as,
\begin{align}\label{eq:aux_loss_mpnn}
    \mathcal{L} =& \underbrace{\sum_{u \in \mathbb{V}_{\tau}} l(\mathrm{f}(\mathcal{G}^{sub}_{u}; \boldsymbol{\theta}), \mathbf{y}_u)}_{\text{loss of the current task $\mathcal{L}_{\tau}$}} \nonumber\\
    &+ \lambda \underbrace{\sum_{\bar{\mathcal{G}}^{sub}_{v} \in \mathcal{SSM}} l(\mathrm{f}(\bar{\mathcal{G}}^{sub}_{v}; \boldsymbol{\theta}), \mathbf{y}_v)}_{\text{auxiliary loss $\mathcal{L}_{aux}$}}.
\end{align}

Based on SSM, a Ricci-curvature based Subgraph Episodic Memory (SEM)~\cite{zhang2023ricci} is further constructed to sparsify the computation subgraphs based on the information bottleneck of each edge. The information bottleneck of is measured with its Ricci-curvature, which is a theoretically justified metric for evaluating the smoothness of information flow via each edge~\cite{topping2021understanding}.
The Ricci-curvature of an edge $e_{uv}$ is defined as:
\begin{align}\label{eq:ric}
    Ric(u,v):=&\frac{1}{d_u}+\frac{1}{d_v}-2+2\frac{|\#_\triangle(u,v)|}{\max\{d_u,d_v\}} + \frac{|\#_\triangle(u,v)|}{\min\{d_u,d_v\}}&& \nonumber\\ 
    &+ \frac{\gamma_{\max}^{-1}}{\max\{d_u,d_v\}}(|\#_{\square}^u(u,v)|+|\#_{\square}^v(u,v)|),&&
\end{align}
    where $d_u$ is the degree of node $u$, $\#_\triangle(u,v):= \mathcal{N}^1(u) \cap \mathcal{N}^1(v)$ denotes the number of triangles formed with $u$, $v$ and one of their common 1-hop neighbors. $\#_{\square}^v(u,v):= \{w  \in \mathcal{N}^1(u) | w \neq v, \exists q \in (\mathcal{N}^1(w)\cap \mathcal{N}^1(v)) \backslash  \mathcal{N}^1(u) \}$ denotes the number of 4-cycles. Both of them count the higher order paths for information flow. $\gamma_{\max}=\max\big\{\max_{p\in \#_{\square}^u(u,v)}\{|\mathcal{N}^1(p)\cap \#_{\square}^v(u,v) \backslash \mathcal{N}^1(u) |-1 \},\max_{q\in\#_{\square}^v(u,v)}\{|\mathcal{N}^1(q) \cap \#_{\square}^u(u,v) \backslash \mathcal{N}^1(v) |-1 \} \big\} $ is the maximal number of 4-cycles at edge $e_{u,v}$ via a common node, and serves as a normalization term.
However, since Ricci-curvature has a high computational complexity, a random walk based surrogate is introduced to indicate the relative magnitude of the Ricci-curvature of each edge. Finally, the edges are selected by the sampling from the random walk distribution.

SSM is a model agnostic method that can be implemented with any GNN. Another recently proposed memory based CGL approach, PDGNNs-TEM~\cite{zhang2024topology}, formulates a framework to store the complete topological information of each computation subgraph with a single embedding vector. Specifically, PDGNNs-TEM consists of a general GNN framework named as Parameter Decoupled Graph Neural Networks (PDGNNs) and a memory buffer named as Topology-aware Embedding Memory (TEM). As analyzed by \cite{zhang2022sparsified}, the key challenge in storing complete topological information of graph data is the memory explosion problem, \textit{i.e.} the size of the computation subgraph grows exponentially with the number of hops. Therefore, PDGNNs is proposed to decouple the trainable parameters from the input computation subgraph, so that the nodes in the computation subgraph do not interact with the trainable parameters individually, and retraining the model only needs an overall information of each computation subgraph. Specifically, instead of iteratively aggregating the neighboring nodes and conducting node feature transformation, PDGNNs first encode the entire computation subgraph into a topology-aware embedding with a non-parametric function to capture the topological information of the computation subgraph,
\begin{align}\label{eq:sse_func}
    \mathbf{e}_v =\mathrm{f}_{topo} (\mathcal{G}^{sub}_v).
\end{align}
In Equation~\ref{eq:sse_func}, $\mathbf{e}_v$ denotes the topology-aware embedding (TE) for node $v$, $\mathcal{G}^{sub}_v$ is the computation subgraph of node $v$, and $\mathrm{f}_{topo}(\cdot)$ is the non-parametric function that can be instantiated with different forms. In the paper, both linear formulation and non-linear formulation of $\mathrm{f}_{topo}(\cdot)$ are explored. After that, the obtained TEs are then fed into a trainable function to generate the final output prediction,
\begin{align}
    &\mathbf{\hat{y}}_v = \mathrm{f}_{out}(\mathbf{e}_v; \boldsymbol{\theta}).
\end{align}
With the PDGNNs framework, for retraining the model with a certain node $v$, the entire computation subgraph $\mathcal{G}^{sub}_v$ is no longer needed. Instead, only the TE of $v$ needs to be stored for re-training the trainable function $\mathrm{f}_{out}(\cdot)$. Accordingly, the Topology-aware Embedding Memory ($\mathcal{TEM}$) is developed to store the selected TEs for each task,
\begin{align}\label{eq:TEM}
    \mathcal{TEM}=\mathcal{TEM} \; \bigcup \; \mathrm{sampler}(\{\mathbf{e}_v \mid v\in \mathbb{V}_{\tau}\}, n),
\end{align}
where $\mathrm{sampler}(\cdot,\cdot)$ is the chosen sampling strategy for populating the memory buffer, $n$ is the budget for storage, and $\mathbb{V}_{\tau}\}$ denotes the node set of the current task. Compared to ER-GNN and SSM, TEM can maintain complete topological information for each node with only one emebdding vector, therefore is highly efficient in terms of both space complexity and computation complexity during memory replay. But TEM has to be accompanied with a GNN following the PDGNNs framework, while ER-GNN and SSM can be implemented with any GNN.

As mentioned before in Section~\ref{sec:regularization based methods}, the inter-task edge connections would alter the neighborhood of the nodes in the previous tasks and cause concept shift, while the methods mentioned above do not explicitly consider this problem. Targeting this challenge, Structure-Evolution-Aware Experience Replay (SEA-ER)~\cite{su2023towards_seaer} is proposed to explicitly consider the evolution of the graph structure when populating the memory buffer. Denoting the memory buffer for the $j$-th task as $P_j$, the objective for populating the buffer is to maximize the structural similarity between the selected samples and the rest nodes,
\begin{align}
    \min_{P_j \subset \mathcal{V}_j} \max_{u\in \mathcal{V}_j\backslash P_j} \min_{v \in P_j} \mathrm{d}_{\mathrm{spd}(u,v)}, \quad s.t.|P_j| = b,
\end{align}
where $\mathrm{d}_{\mathrm{spd}(u,v)}$ denotes the shortest path distance between node $u$ and $v$, $b$ is the memory budget for task $j$, and $\mathcal{V}_j$ is the node set of the task $j$. 
Since the evolution of the graph structure may also influence the distribution of the data in previous tasks, besides the sampling strategy, an importance reweighting technique is also designed in SEA-ER to rescale the contribution of different nodes. Specifically, the objective function with the node reweighting for learning on task $i$ is formulated as,
\begin{align}
    \mathcal{L} = \frac{1}{|\mathcal{V}^{train}_i|} \sum_{v\in \mathcal{V}^{train}_i} \mathcal{L} + \sum_{P_j \in \mathcal{P}_i}  \frac{1}{|P_j|} \sum_{v\in P_j} \beta_v \mathcal{L}(v),
\end{align}
where $\mathcal{P}_i = \{P_1, ..., P_{i-1}\}$ denotes the memory buffer during learning task $i$, which contains representative data selected from all previous tasks. $\beta_v$ is the reweighting coefficient on each node $v$, and is obtained by kernel mean matching (KMM)~\cite{gretton2009covariate}, which can be mathematically formulated as the solution of the following problem,
\begin{align}
    \min_{\beta} || \sum_{P_j \in \mathcal{P}_i} \sum_{v \in P_j} \beta_v \phi(h_v) - \phi(h_v') ||^2, s.t. B_l \leq \beta_v \leq B_u,
\end{align}
where $h_v$ and $h_v'$ are node representations generated by the GNN model based on graph from task $i$ and $i-1$, and $\phi(\cdot)$ is the function that maps the node representations into the reproducing Hilbert space associated by the adopted kernel. Besides, $B_l$ and $B_u$ are the lower bound and upper bound on $\beta$ to make sure that the weight is suitable for the majority of the nodes. With the formulation above, the reweighting procedure is actually minimizing the distance between the node representations on new and old tasks by finding the most appropriate weight values. 

Similar to SEA-ER, Incremental Graph Neural Networks (IncreGNN)~\cite{wei2022incregnn} also considers the influence of inter-task edges on nodes in the previous nodes. To tackle this challenge, IncreGNN not only proposes an approach to evaluate the node importance based on both their influence to the new nodes and their influence in the original graph, but also adopts a regularization technique to further boost the performance. Specifically, the newly incoming graph data is denoted as the base change group. Based on the base change group, the 1-hop neighbors of the base change group in the previous task graph is defined as the first order change group, the 1-hop neighbors of the first order change group is defined as the second order change group, and so forth. Since the influence of the new nodes on the previous nodes decreases with the increase of the order of the changing group, the ratio of the budget for node sampling at different order of change group is designed as,
\begin{align}
    \frac{1}{i} / (1+\frac{1}{2}, ..., +\frac{1}{K}),
\end{align}
where $i$ is the number of order that is being sampled, $K$ is the total number of change groups.
After defining the importance of the groups of nodes with different distance to the new nodes, the importance of an individual node $v$ is evaluated by the personalized PageRank algorithm run on the previous task graph $\mathcal{G}^{t-1}$, and is denoted as $\pi_v$. Then, denoting the number of nodes to sample at $k$-order of change group as $n_k$, the nodes at each order with higher personalized PageRank values are selected as the ones with more correlation with the new nodes,
\begin{align}
    I(\mathcal{G}^t) = \cup_{k=1}^K \{ v_i | \pi_{v_i} > \pi_{v_j} , i \in [1,n_k], j\notin [1,n_k] \}.
\end{align}
The above process only select the old nodes affected by the new nodes. To also maintain the knowledge learnt from the old nodes without inter-task edges to new nodes (unaffected nodes), the next step is to evaluate the importance of each unaffected node for sampling and storage. Specifically, K-means clustering is first conducted to divide the unaffected nodes into $K$ clusters. Then, the node importance within each cluster is determined by the node degree, and the nodes with the highest degrees are selected and stored in the memory buffer.
Finally, IncreGNN also adopts the Memory Aware Synapse (MAS)~\cite{aljundi2018memory} as the regularization on the model parameters to further enhance the capability to retain learnt knowledge.

Different from the approaches above, which retrain the same model structure with buffered data, \cite{rakaraddi2022reinforced} proposes to also dynamically adjust the model to accommodate new tasks. The proposed model consists of two parts including a Reinforcement Learning based Controller (RLC) that decides the addition and pruning of the model structure,  and a Child Network (CN) which is a evolvable GNN backbone controlled by the RLC. The RLC is based on reinforcement learning, which aims to learn an optimal policy $\pi(a_t|s_t))$ to choose the most appropriate action $a_t$ to modify the model based on the model state at the $t$-the task, so that the expected summation of the future rewards $R_t = \sum_{k=0}^{\infty} \gamma^k_{t+k+1}$ is maximized. 
The decaying factor $\gamma \in (0,1]$ serves to gradually shrink the contribution of the expected rewards at distant time steps. Accordingly, the Q-function of RLC for taking an action $a_t$ with a model state $s_t$ is,
\begin{align}
    Q_{\pi}(s_t,a_t) = \mathrm{E}_{\pi}[R_t|S=s_t,A=a_t].
\end{align}
The RLC in \cite{rakaraddi2022reinforced} is designed as an LSTM, which outputs a set of $m$ actions $a_{1:m}$ for adjusting the number of features in each of the $m$ layers of the CN. The search space for the action is defined as the addition and deletion operation of the CN hidden layers. When RLC is properly trained, the optimal actions for adjusting the CN is formulated as,
\begin{align}
    \hat{a}_{1:m} = \arg\max_{a \in \mathcal{S}} \mathbb{E}[R(a_{1:m},s_t)].
\end{align}
Besides the RLC to optimize the structure of CN, to further enhance the continual learning capability of the model, the Dark Experience Replay~\cite{buzzega2020dark} is also adopted after being extended to graph data. Specifically, besides storing the data and the labels of the selected examples, their logits output by the CN is also buffered. During replay, the model retraining loss would accordingly consists of two parts including the classification loss calculated with the buffered labels and data, as well as the knowledge distillation loss calculated based on the stored logits. The total loss including the loss for learning the new task is then formulated as,
\begin{align}
    \mathcal{L} = \mathcal{L}_t + \alpha || \mathrm{f}_t(\mathcal{X}^i)-l_t ||_2 + \beta \mathcal{L}_{cls}\big( \mathrm{f}_t(\mathcal{X}^i, \mathcal{Y}^i) \big),
\end{align}
where $\mathrm{f}_t(\cdot)$ denotes the current model, $l_i$ is the logits stored from the $i$-the task, and $\mathcal{X}^i$ and $\mathcal{Y}^i$ are the data and labels stored from task $i$.

Besides the models that store real data from learnt tasks, in Streaming Graph Neural Networks via Generative Replay (SGNN-ER), an auxiliary generative model is adopted to retrain the GNN with generated fake historical data. Overall, the generative model would generate the neighborhood (in the form of a random walk with restart sequence) based on the GAN framework. Denoting the generator and discriminator parameterized by $\mathbf{\phi}^t$ and $\varphi^t$ at task $t$ as $\mathrm{G}_{\mathbf{\phi}^t}$ and $\mathrm{D}_{\varphi^t}$, the learning objective can be formulated as a confrontation between the generator and the discriminator,
\begin{align}
    \min_{\mathbf{\phi}^t} \max_{\varphi^t} \mathrm{V}(\mathrm{G}_{\phi^t},\mathrm{D}_{\varphi^t}) & =  \mathrm{E}_{v \sim p_{data}(v)}[\log \mathrm{D}_{\mathbf{\phi}^t}(v)] \\ &+ \mathrm{E}_{z \sim p_z(z)} [\log (1-\mathrm{D}_{\varphi^t} (\mathrm{G}_{\mathbf{\phi}^t}(z)) )] \nonumber.
\end{align}
In SGNN-ER, the generated data is used for knowledge distillation between the old and new models to avoid forgetting, and the total loss can be formulated as
\begin{align}
   \nonumber \mathcal{L} & = r \mathrm{E}_{v \sim \mathcal{G}_A^t} [l(\mathrm{F}_{\boldsymbol{\theta^t}}(v)),y_v] \\ & + (1-r) \mathrm{E}_{v' \sim G_{\phi}^{t-1}}[l(\mathrm{F}_{\boldsymbol{\theta}^t}(v'),\mathrm{F}_{\boldsymbol{\theta}^{t-1}}(v'))], 
\end{align}
where $\mathcal{G}_A^t$ is the affected part at task $t$, including the newly added subgraph and the previous nodes connected to the newly added subgraph. 

Originally, the Generative Adversarial Network (GAN) framework is used for generating independent data. However, to generate the representation of a single node, GNNs would also take its multi-hop neighborhood with complex topological dependencies as input. Therefore, SGNN-ER proposes to generate random walk with restart (RWR) sequences, which are then converted into neighborhood subgraphs based on the generated edge connectivity.
Considering the neighborhood of some previous nodes is changed by the newly added nodes, to avoid enforcing the patterns that already disappear because of the neighborhood change, SGNN-ER proposes a forgetting mechanism to filter out the nodes that are significantly affected by the new nodes during memory replay. The mechanism consists of two steps. First, a set of significantly affected nodes $\mathcal{V}^t_C$ is first detected based on the change in their representations when the graph grows from $\mathcal{G}^{t-1}$ to $\mathcal{G}^{t}$,
\begin{align}
    \mathcal{V}^t_C = \{ v| ||\mathrm{F}_{\boldsymbol{\theta}^{t-1}}(v, \mathcal{G}^t) - \mathrm{F}_{\boldsymbol{\theta}^{t-1}}(v,\mathcal{G}^{t-1})||>\delta \},
\end{align}
where $\delta$ is the threshold on the change of representations.
Since $\mathcal{V}^t_C$ denotes the severely affected nodes, among the generated fake nodes for replay, the ones with high similarity to the nodes in $\mathcal{V}^t_C$ are rejected with high probability, which is formulated as,
\begin{align}
    \mathrm{p}_{reject}(v) = \max(\mathrm{p}_{sim}(v,u), u \in \mathcal{V}^t_C) \times p_r, 
\end{align}
where $p_r$ is a hyperparameter controlling the total number of nodes to delete, and $p_{sim}(\cdot,\cdot)$ is the similarity function for measuring the similarity between two nodes,
\begin{align}
    \mathrm{p}_{sim}(v,u) = \sigma(-|| \mathrm{F}_{\boldsymbol{\theta}^{t-1}}(v,\mathcal{G}^{t-1}) - \mathrm{F}_{\boldsymbol{\theta}}(u, \mathcal{G}^{t-1}) || ).
\end{align}

Another related work in this branch is the Condense and Train (CaT) framework~\cite{liu2023cat}. Different from the works that generate fake data for previous tasks, to balance the contribution of different tasks, CaT is designed to train the model only on the generated balanced graphs. Specifically, when a new task ($\mathcal{G}_{\tau}$) comes, CaT will first condense it based on the graph node embeddings generated by the same GNN encoder,
\begin{align}
    \Tilde{\mathcal{G}}_{\tau}=\arg\min_{\Tilde{\mathcal{G}}_{\tau}^*} \mathrm{Dist}(\mathrm{GNN}(\mathcal{G}_{\tau};\theta),\mathrm{GNN}(\Tilde{\mathcal{G}}_{\tau}^*;\theta)).
\end{align}
Denoting the node embeddings $\mathrm{GNN}(\mathcal{G}_{\tau};\theta)$ and $\mathrm{GNN}(\Tilde{\mathcal{G}}_{\tau}^*;\theta)$ as $\mathbf{E}_{\tau}$ and $\mathbf{E}_{\tau}^*$, the distant function $\mathrm{Dist}(\cdot,\cdot)$ is formulated as a weighted summation over the mean discrepancy between the node embeddings of each class in the two graphs,
\begin{align}
\label{eq:CaT_dist}
    \mathrm{Dist}(\mathbf{E}_{\tau},\mathbf{E}_{\tau}^*) = \sum_{c\in \mathcal{C}_{\tau}} \frac{|\mathbf{E}_{\tau,c}|}{|\mathbf{E}_{\tau}|} \cdot ||\mathrm{Mean}(\mathbf{E}_{\tau,c})-\mathrm{Mean}(\mathbf{E}_{\tau,c}^*)||.
\end{align}
In equation~\ref{eq:CaT_dist} above, $\frac{|\mathbf{E}_{\tau,c}|}{|\mathbf{E}_{\tau}|}$ denotes the ratio of data belonging to class $c$ in the entire set of data, and $\mathrm{Mean}(\cdot)$ refer to the mean of the embedding vectors. Next, to faithfully reflect the distance between the graphs in different embedding spaces, $\mathrm{GNN}(\cdot;\theta)$ in CaT is initialized with different random parameters $\theta$, and the distance is averaged over all random parameter initialization.
The obtained condensed graph of the current task is updated into a memory,
\begin{align}
    \mathcal{M}_{\tau} = \mathcal{M}_{\tau-1} \cup \{\Tilde{\mathcal{G}}_{\tau}\},
\end{align}
where $\mathcal{M}_{\tau}$ denotes the memory buffer containing all condensed graphs from the observed $\tau$ tasks. Finally, the model is trained solely on $\mathcal{M}_{\tau}$ for a balanced training.

The methods mentioned above are general CGL methods that are applicable to any graph data. For certain graph related tasks, due to the special properties of the graph data, some specialized techniques may be developed. For example, specially designed for incremental learning in graph based recommender system, Structure Aware Experience Replay (SAER)~\cite{ahrabian2021structure} samples and stores representative user-item pairs based on reservoir sampling, so that the previously learnt user behaviour patterns can be maintained when learning new tasks. Specifically, the reservoir is a set of previously observed interactions denoted as $\mathcal{H}$. When a new set of interactions, \textit{i.e.} $\mathcal{H}'$, is given, a subset $\hat{\mathcal{H}}$ will firstly be sampled from $\mathcal{H}$ according to an adopted sampling strategy. Then the model will be trained with $\mathcal{H}'\cup \hat{\mathcal{H}}$. In \cite{ahrabian2021structure}, two strategies are adopted. The first one is the uniform sampling,
\begin{align}
    \hat{\mathcal{H}} = \{(u,i)\sim U(\mathcal{H})\} \quad s.t \quad |\hat{\mathcal{H}}|=\gamma |\mathcal{H}|,
\end{align}
where $\gamma$ is a hyperparameter controlling the size of the sampled set. The uniform sampling serves as a naive baseline, and a more complex strategy is also adopted to balance the sampled interactions according to their degree, the probability for sampling an interaction $(u,i)$ is,
\begin{align}
    \mathrm{P}(u,i) = \frac{1/d_u^T}{\sum_{(\hat{u},\hat{i}) \in \mathcal{H}}1/d_{\hat{u}}^T},
\end{align}
in which $d_u$ is the degree of the user $u$, and the temperature $T$ is a hyperparameter for regulating the smoothness of the probability distribution.
The pairwise data structure is commonly adopted by recommender systems, but is not available for general graph data. Therefore this technique is not generally applicable.

Currently, memory replay based methods are among the most effective approaches and are easy to implement. But it requires extra space to store representative data.

\subsection{Parameter isolation based methods}

\begin{figure}
    \centering
    \includegraphics[width=0.5\textwidth]{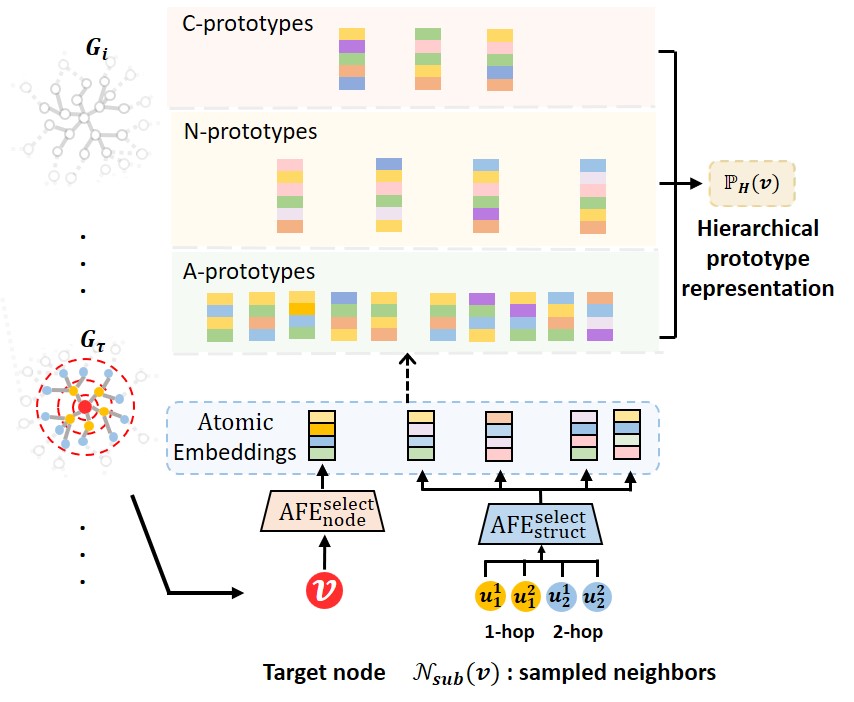}
    \caption{Pipelines of HPNs~\cite{9808404} stores sparsified computation subgraphs.} 
    \label{fig:HPNs}
\end{figure}

The final category, parameter-isolation based methods, protects the model's performance on previous tasks by entirely or partially isolating the model parameters for different tasks. Existing parameter isolation based CGL approaches are scarce, and one representative method is Hierarchical Prototype Networks (HPNs) that propose to dynamically increment feature extractors and prototypes for accommodating new patterns. Specifically, HPNs consist of a set of Atomic Feature Extractors (AFEs) for extracting basic features from the given data, and three levels of hierarchical prototypes for storing learnt patterns. Given an input node, the AFEs will first extract basic features based on both the node attributes and its neighborhood relationship. Accordingly, the AFEs consist of two parts. One set of AFEs, denoted as $\mathrm{AFE}_{\textrm{node}}$, serve to generating atomic node embeddings based on node attributes. The other set, denoted as $\mathrm{AFE}_{\textrm{struct}}$, serve to capture the topological structure based on the neighborhood of the given node, and generate atomic structure embeddings accordingly. The obtained atomic embeddings correspond to the most basic features, and will then be matched to the existing atom-level prototypes (A-prototypes) according to their cosine similarity.
\begin{align}\label{eq:ea_sim}
    \mathrm{Sim}_{{E\rightarrow A}}(v) = \{\frac{\mathbf{e}_i^T\mathbf{p}}{||\mathbf{e}_i||_2 ||\mathbf{p}||_2} | \mathbf{e}_i\in \mathbb{E}^{\textrm{select}}_A(v), \mathbf{p} \in \mathbb{P}_A \},
\end{align}
where $\mathbb{E}^{\textrm{select}}_A(v)$ contains the atomic embeddings of the node $v$, and $\mathbb{P}_A$ is the set of A-prototypes. When the cosine distance between an A-prototype and an embedding is smaller than a threshold $t_A$, the A-prototype and the embedding is regarded as matched. If an embedding cannot be matched to any existing prototype, then it is regarded as new knowledge and will be used to initialize a new prototype. After A-prototype matching, the matched A-prototype will be further embedded and matched to higher level prototypes including the node-level prototypes (N-prototypes) and the class-level prototypes (C-prototypes). Each node will be matched to one N-prototype denoting the overall information of the entire node, and one C-prototype representing the properties of the class it belongs to.  The union of the selected prototypes at different level are the final representation of the node $v$ (Figure~\ref{fig:HPNs}). Besides a novel model, The authors of \cite{9808404} also propose a theoretical framework to analyze the continual learning capability of a model and justify that the proposed HPNs can theoretically eliminate the forgetting problem when the hyperparameters are properly set.

Different from the majority of CGL works that solely focus on the growth of the graph, Parameter Isolation GNN (PI-GNN)~\cite{zhang2023continual} studies the CGL problem in the dynamic graph setting, in which both node addition and deletion are considered. Specifically, when the training of a $k$-layer GNN proceeds from task $\tau-1$ to $\tau$, the graph evolves from $\mathcal{G}_{\tau-1}$ to $\mathcal{G}_{\tau}$. In this process, $\mathcal{G}_{\tau-1}$ can be divided into two parts: $\mathcal{G}_{\tau-1}^{unstable}$ and $\mathcal{G}_{\tau-1}^{stable}$. $\mathcal{G}_{\tau-1}^{unstable}$ consists of the nodes whose $k$-hop neighborhood is altered by the graph evolution (\textit{e.g.}, being connected to new nodes), while $\mathcal{G}_{\tau-1}^{stable}$ refers to the remaining part. Then, the changed part of the graph from $\mathcal{G}_{\tau-1}$ to $\mathcal{G}_{\tau}$ is denoted as $\Delta\mathcal{G}_{\tau}$.
When learning on a new task $\tau$, the main idea of PI-GNN is to first preserve the stable knowledge in $\mathcal{G}_{\tau-1}^{stable}$, and then allocate new separated parameters to accommodate the new knowledge. The stable knowledge preservation is formulated as only optimizing the parameters on $\mathcal{G}_{\tau-1}^{stable}$,
\begin{align}
    &\min_{\theta_{\tau-1}} \mathcal{L}\big( \mathrm{f}(\mathcal{G}_{\tau-1}^{stable};\theta_{\tau-1}) \big)\\
    \textit{i.e.} & \min_{\theta_{\tau-1}} \mathcal{L}\big( \mathrm{f}(\mathcal{G}_{\tau-1};\theta_{\tau-1}) \big) - \mathcal{L}\big( \mathrm{f}(\mathcal{G}_{\tau-1}^{unstable};\theta_{\tau-1}) \big).
\end{align}
Since computing $\mathcal{L}\big( \mathrm{f}(\mathcal{G}_{\tau-1};\theta_{\tau-1}) \big)$ is time consuming, and the access to previous data may not always be available, $\mathcal{G}_{\tau-1}$ is approximated with a set of representative data $\mathcal{G}_{\tau-1}^{memory}$ sampled from $\mathcal{L}\big( \mathrm{f}(\mathcal{G}_{\tau-1};\theta_{\tau-1}) \big)$, \textit{i.e.},
\begin{align}
    \theta_{\tau-1}^{stable} = \arg\min_{\theta_{\tau-1}} \mathcal{L}\big( \mathrm{f}(\mathcal{G}_{\tau-1}^{memory};\theta_{\tau-1}) \big) - \\ \nonumber \beta \mathcal{L}\big( \mathrm{f}(\mathcal{G}_{\tau-1}^{unstable};\theta_{\tau-1}) \big).
\end{align}
Since the memory data $\mathcal{G}_{\tau-1}^{memory}$ has a different size compared to the original graph $\mathcal{G}_{\tau-1}$, the hyperparameter $\beta$ is introduced for the size imbalance issue.
$\theta_{\tau-1}^{stable}$ is fixed after encoding the previous knowledge, and new parameters $\theta_{\tau}^{new}$ are allocated for encoding the new knowledge in $\Delta\mathcal{G}_{\tau}$ and $\mathcal{G}_{\tau-1}^{memory}$. Finally, the parameters obtained after learning task $\tau$ is $\theta_{\tau} = \theta_{\tau-1}^{stable} \oplus \theta_{\tau}^{new}$.

\subsection{Application of Traditional Continual Learning Techniques}
Although not specialized for graph data, some of the traditional CL techniques for Euclidean data are both data- and model-agnostic, therefore are applicable to CGL. For example, regularization based methods like Elastic Weight Consolidation (EWC) \cite{kirkpatrick2017overcoming} and Memory Aware Synapses (MAS) \cite{aljundi2018memory} that only add regularization terms to the model weights, memory replay based models like Gradient Episodic Memory (GEM)~\cite{lopez2017gradient} that stores the gradients of the losses on previous tasks and clip the current gradients to avoid increasing the previous task losses, and parameter-isolation based methods like Progressive Neural Networks~\cite{rusu2016progressive} and Supermasks in Superposition (SupSup)~\cite{wortsman2020supermasks} that entirely or partially separate parameters for different tasks. The key characteristic of these works that makes them applicable to CGL is that they are agnostic of the model structure and the data structure. While the other methods not satisfying this criterion are not directly applicable to CGL. For example, most memory replay based methods would directly store representative data from previous tasks. However, on a growing graph, since GNNs generate the representation of one datum (node) not only based on itself but also on its multi-hop neighbors (computation subgraph), and different GNNs may adopt different computation subgraph constructions, it is unclear which part of the graph should be stored.

Targeting the inapplicability of some of the traditional CL methods, Feature Graph Network (FCN) \cite{wang2022lifelong} is proposed to accommodate the data by transforming each node into an independent feature graph, so that the data structure is no longer an obstacle for applying traditional continual learning techniques. When generating the feature graph, the topological information is also implicitly encoded. FCN provides a general approach to enable traditional CL techniques in any graph data. But for some special graphs, like knowledge graphs, since the data storage format is more tidy, traditional CL techniques are applicable with little modification. For example, considering that the knowledge graph in robotics applications needs to be frequently updated with new information, \cite{daruna2021continual} extends several continual learning techniques including Progressive Neural Networks \cite{rusu2016progressive}, Copy weight Re-init Deep Generative Replay~\cite{shin2017continual} to knowledge graph embedding for avoiding learning all the concepts afresh.  We denote the ensemble of methods proposed by \cite{daruna2021continual} as Adapted Continual Learning Methods (ACLM) for convenience.


\section{Evaluation of CGL Models}\label{sec:evaluation}
\begin{figure}
    \centering
    \includegraphics[width=0.5\textwidth]{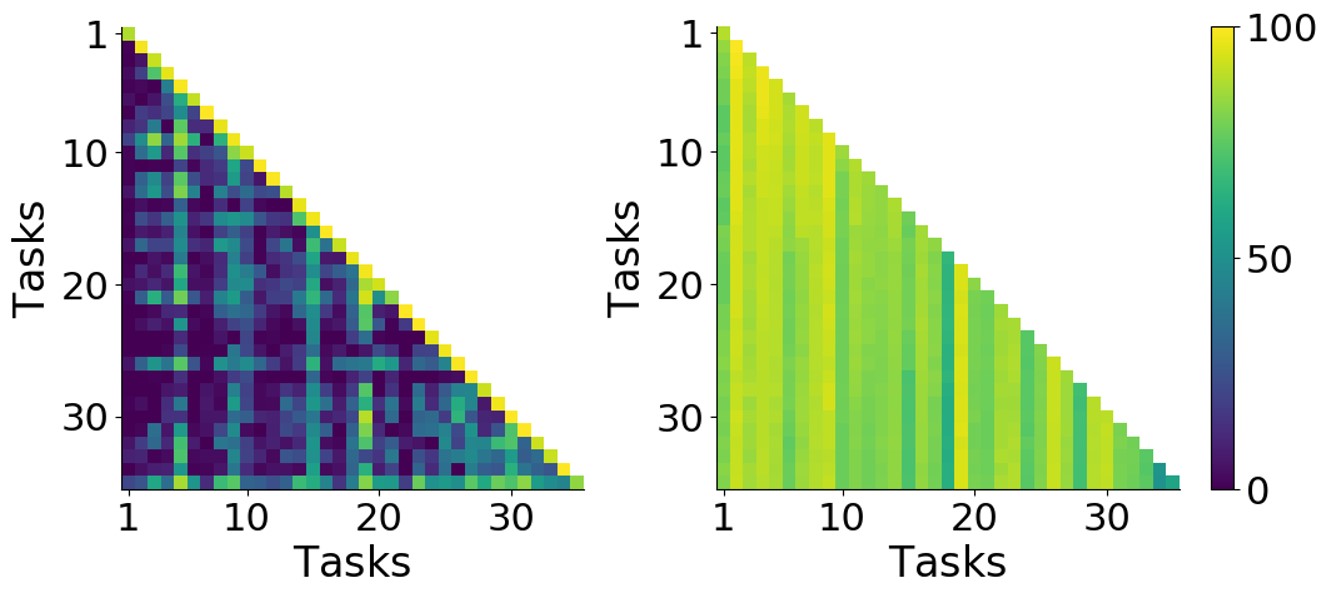}
    \caption{An example of two visualized performance matrices provided in CGLB~\cite{zhang2022cglb}} 
    \label{fig:heatmap}
\end{figure}

Different from standard learning setting that is only concerned with one task, the evaluation of continual learning models have to consider the model's performance on all learnt tasks. Therefore, the most through approach to evaluate a continual learning model is to show its performance on each previous task after learning each new task. Formally, this result could be represented as a performance matrix $\mathrm{M}^{p}\in \mathbb{R}^{T \times T}$ \cite{zhang2022cglb}. Each entry $\mathrm{M}^{p}_{i,j}$ is the model's performance on the $j$-th task after the model has learnt the first $i$ tasks. The performance matrix contains the raw performance during the entire learning process, which can be visualized as a heatmap as shown in Figure~\ref{fig:heatmap}.
Based on the performance matrix, multiple different metrics are adopted by the researchers. For example, considering that the performance matrix is inconvenient for performance comparisons across different methods, the average performance (AP) and average forgetting (AF) can be used~\cite{zhang2022cglb}. These two metrics are widely adopted in CGL as well as traditional continual learning works, while the names may be different in different works. For example, they are named as Average Accuracy (ACC) and Backward Transfer (BWT) in \cite{chaudhry2018riemannian,lopez2017gradient}, Average Performance (AP) and Average Forgetting (AF) in \cite{zhang2022cglb,zhou2021overcoming} and Performance Mean (PM) and Forgetting Mean (FM) in \cite{liu2021overcoming}. Broadly speaking, the entries of the performance matrix may not only be accuracy and can also be other metrics like F1 scores. Therefore, we follow the names (AP and AF) in CGLB~\cite{zhang2022cglb}. Given a task sequence of length $T$, AP is calculated as the average of the performance on all previous tasks after learning the entire task sequence $\frac{\sum_{j=1}^T \mathrm{M}^{p}_{T,j}}{T}$. The AF is calculated as the average of the forgetting (performance decrease). The forgetting has two slightly different versions in different works. First, the forgetting of a task $j$ after learning the final task $T$ could be defined as $f_j = \mathrm{M}^{p}_{T,j}-\mathrm{M}^{p}_{j,j}$, which indicates how much the performance decreases from the time when the model has just learnt task $j$ to the time when the model has learnt all tasks~\cite{lopez2017gradient,zhang2022cglb}. Another definition compares the final performance with the best performance ever obtained when the model is trained from task $1$ to task $T$, \textit{i.e.} $f_j = \max_{i =1,...,T-1} \mathrm{M}^{p}_{i,j}- \mathrm{M}^{p}_{T,j}$. Different from the first definition, this definition compare the final performance on task $j$ with the best performance ever obtained on task $j$ before learning on the final task $T$. Another difference between these two definitions is that the sign of the computation. With the definition of the forgetting, the average forgetting (AF) can be naturally defined as $\frac{\sum_{j=1}^{T-1} f_j}{T-1} $. Further, to reflect the learning dynamics when learning sequentially on the tasks, the AP and AF after learning each task can be calculated, \textit{i.e.} 
$ \Big\{\frac{\sum_{j=1}^i \mathrm{M}^{p}_{i,j}}{i} | i=1,...,T \Big\}$ and $\Big\{\frac{\sum_{j=1}^{i-1} \mathrm{M}^{p}_{i,j}-\mathrm{M}^{p}_{j,j}}{i-1} | i=2,...,T \Big\}$.
Besides AP and AF, researchers are also interested in the forward transfer during learning on the task sequence, and a forward transfer is also defined~\cite{lopez2017gradient} as, $\frac{\sum_{j=2}^{T} \mathrm{M}^{p}_{j-1,j}- \bar{b_j}}{T-1} $, where $\bar{b_j}$ denotes the performance on task $j$ with random model initialization. Accordingly, $\mathrm{M}^{p}_{j-1,j}- \bar{b_j}$ reflects how does the learning from task 1 to task $j-1$ benefit the performance on task $j$, \textit{i.e.} forward transfer.

\section{CGL Benchemarks \& Datasets}\label{sec:benchmark}

Since CGL is a newly emerging area, different works adopt different datasets and conduct experiments under different settings, a fair comparison across different methods becomes difficult, which greatly impedes the development of the area. Accordingly, several benchmark works are proposed to provide a consistent setting and dataset protocol for developing and comparing CGL techniques. In 2021, a benchmark for CGL with graph-level tasks is introduced~\cite{carta2022catastrophic}. The benchmark is constructed based on three datasets: MNIST, CIFAR10, and OGBG-PPA. MNIST and CIFAR10 are image datasets that are originally used for computer vision. However, following the process in \cite{dwivedi2020benchmarking}, the images are converted to graphs. Given an image, with the SLIC algorithm~\cite{achanta2012slic}, the pixels within a small local area with homogeneous intensity are defined as a super-pixel, corresponding to a node in the converted graph. The topology (graph edges) is constructed as the k-nearest neighbor adjacency matrix. For the task construction, both MNIST and CIFAR10 contains 10 classes in total, and the classes are divided into five 2-class tasks.
The OGBG-PPA~\cite{hu2020ogb} dataset contains a set of protein association neighborhoods (subgraphs) extracted from the large protein protein interaction (PPI) network. OGBG-PPA consists of 37 classes, each of which corresponds to a taxonomy group. The 37 groups are also split into 5 tasks. The first task contains 17 classes while the other four classes contain 5 classes. 
Constructing the tasks by splitting the classes of the data is a common practice for CGL, because the data in different classes generally follow different distributions, which easily cause catastrophic forgetting and is ideal for CGL tasks.
Besides the benchmark task construction, experiments of 4 baselines including Naive, EWC, Replay, and LwF are conducted under the class-IL scenario. The accompanied code of this benchmark is provided through \url{https://github.com/diningphil/continual_learning_for_graphs}.

In 2022, the Continual Graph Learning Benchmark (CGLB)~\cite{zhang2022cglb} is proposed, which is a more comprehensive benchmark covering 4 node-level datasets and 3 graph-level datasets under both task-IL and class-IL scenarios. The node-level datasets include CoraFull-CL, Arxiv-CL, Reddit-CL, and Products-CL. Both CoraFull-CL and Arxiv-CL are constructed based on citation networks provided in the public datasets CoraFull~\cite{mccallum2000automating} and OGB-Arxiv~\cite{hu2020ogb}. Reddit-CL is constructed from social network consisting of Reddit posts~\cite{hamilton2017inductive}. Products-CL is constructed from the OGB-Products dataset~\cite{hu2020ogb}, which is based on the Amazon product co-purchasing network. In CGLB, to maximize the number of tasks and increase the continual learning difficulty, in CGLB, all of these node-level tasks are split into 2-class tasks. For node-level CGL tasks, the growth of the graph would bring in inter-task edges connecting the nodes in previous and new tasks, and cause concept drift to the previous nodes connected by the inter-task edges. Accordingly, whether to preserve the inter-task edges when increasing the tasks is studied separately in CGLB. All of these node-level benchmark datasets support both task-IL and class-IL learning scenarios. On the node-level benchmark datasets, preliminary experimental results of multiple baseline methods are also provided under different settings. 
For graph-level CGL tasks, three benchmark datasets are constructed in CGLB including SIDER-tIL, Tox21-tIL, and Aromaticity-CL. SIDER-tIL is constructed from the SIDER dataset\cite{wu2018moleculenet}, which contains 1,427 drugs falling into 27 different classes. Since SIDER is a multi-label dataset, it is naturally constructed into a task-IL dataset with 27 tasks. 
Tox21-tIL is constructed from the Tox21 dataset~\cite{wu2018moleculenet}, which is a multi-label dataset containing 8,014 molecules with 12 labels. Similar to SIDER, Tox21 is also constructed into a task-IL dataset with 12 tasks.
Aromaticity-CL is constructed from the PubChem BioAssay Dataset~\cite{xiong2019pushing}, which contains 3,945 molecules falling into 40 classes. Since PubChem BioAssay Dataset is a multi-class dataset, the constructed Aromaticity-CL contains fifteen 2-class tasks, and can be used for both task-IL and class-IL. In Aromaticity-CL, some classes with very few data are removed. The accompanied code of CGLB is provided via \url{https://github.com/QueuQ/CGLB}.

Finally, we list all commonly used datasets in Table~\ref{tab:datasets}. Among these datasets, some are constructed as standard CGL tasks (CoraFull, OGB-Arxiv, Reddit and OGB-Products are included in CGLB~\cite{zhang2022cglb}), while the others may be configured differently in different works. 

\begin{table*}[t] 
\scriptsize
\captionsetup{font=small}
    \caption{Commonly used datasets for CGL.}
    \centering
    \begin{tabular}{lp{6cm}p{8cm}}\toprule
        Dataset & Related works & Link \\\midrule
        CoraFull~\cite{mccallum2000automating} & TWP~\cite{liu2021overcoming},SSM~\cite{zhang2022sparsified},PDGNNs-TEM~\cite{zhang2024topology},RLC-CN~\cite{rakaraddi2022reinforced},CGLB~\cite{zhang2022cglb},SEM~\cite{zhang2023ricci},CaT~\cite{liu2023cat},UGCL~\cite{hoang2023universal}  & \url{github.com/shchur/gnn-benchmark#datasets} \\ \midrule
        OGB-Arxiv~\cite{hu2020ogb} &RieGrace~\cite{sun2023self},SSM~\cite{zhang2022sparsified},PDGNNs-TEM~\cite{zhang2024topology},HPNs~\cite{9808404},CGLB~\cite{zhang2022cglb},SEM~\cite{zhang2023ricci},CaT~\cite{liu2023cat},UGCL~\cite{hoang2023universal} & \url{ogb.stanford.edu/docs/nodeprop/\#ogbn-arxiv} \\ \midrule
        Reddit~\cite{hamilton2017inductive} &TWP~\cite{liu2021overcoming},RieGrace~\cite{sun2023self},DyGRAIN~\cite{kim2022dygrain},ER-GNN~\cite{zhou2021overcoming},SSM~\cite{zhang2022sparsified},PDGNNs-TEM~\cite{zhang2024topology},SEA-ER~\cite{su2023towards_seaer},CGLB~\cite{zhang2022cglb},SEM~\cite{zhang2023ricci},CaT~\cite{liu2023cat},UGCL~\cite{hoang2023universal} & \url{snap.stanford.edu/graphsage/}  \\\midrule
        OGB-Products~\cite{hu2020ogb} &DyGRAIN~\cite{kim2022dygrain},SSM~\cite{zhang2022sparsified},PDGNNs-TEM~\cite{zhang2024topology},HPNs~\cite{9808404},CGLB~\cite{zhang2022cglb},SEM~\cite{zhang2023ricci},CaT~\cite{liu2023cat} & \url{ogb.stanford.edu/docs/nodeprop/\#ogbn-products} \\\midrule
        Cora~\cite{sen2008collective}  & RieGrace~\cite{sun2023self},ER-GNN~\cite{zhou2021overcoming},RLC-CN~\cite{rakaraddi2022reinforced},SGNN-ER~\cite{wang2022streaming},SEA-ER~\cite{su2023towards_seaer},HPNs~\cite{9808404},PI-GNN~\cite{zhang2023continual} & \url{ github.com/tkipf/gcn/tree/master/gcn/data} \\\midrule
        Citeseer~\cite{sen2008collective}  & RieGrace~\cite{sun2023self},ER-GNN~\cite{zhou2021overcoming},RLC-CN~\cite{rakaraddi2022reinforced},SGNN-ER~\cite{wang2022streaming},HPNs~\cite{9808404},PI-GNN~\cite{zhang2023continual} & \url{github.com/tkipf/gcn/tree/master/gcn/data} \\\midrule
        Amazon Computers~\cite{shchur2018pitfalls} & TWP~\cite{liu2021overcoming},RLC-CN~\cite{rakaraddi2022reinforced},PI-GNN~\cite{zhang2023continual} & \url{nijianmo.github.io/amazon/index.html } \\\midrule
        Actor~\cite{tang2009social} & RieGrace~\cite{sun2023self},HPNs~\cite{9808404} & \url{ github.com/graphdml-uiuc-jlu/geom-gcn/tree/master/new_data/film} \\\midrule
        Gowalla~\cite{cho2011friendship} &GraphSAIL~\cite{xu2020graphsail},SAER~\cite{ahrabian2021structure} & \url{snap.stanford.edu/data/loc-gowalla.html} \\
        \bottomrule
    \end{tabular}
    \label{tab:datasets}
\end{table*}


\section{Opportunities, Future Directions and Discussions}\label{sec:Opportunities, Future Directions and Discussions}
Research on CGL has achieved significant advancements, manifesting both in enhanced performance and the exploration of more complex and practical applications. However, as a newly emerging and fast-growing area, there are still multiple challenges to tackle and promising directions to explore. Moreover, CGL is highly relevant to many research areas, and could potentially benefit a vast majority of the graph learning community. In this section, we first introduce several key challenges to be tackled, and then introduce the topics that transcend the boundary of current CGL research and intersect with other areas.

\subsection{Trade-off between Effectiveness and Space Complexity}
Among the three strategies, regularization based methods are the most memory efficient since they only require little extra storage space, while memory-replay based methods and parameter-isolation based methods require significant extra space for storing the representative data or expanded network structure. However, as shown in recent works including benchmark work~\cite{zhang2022cglb} and technical works~\cite{zhang2022sparsified,zhang2024topology,9808404,rebuffi2017icarl}, the regularization based method is not as effective as the other two approaches. The reason is that although the regularization term can effectively restrict the change of the model parameters to prevent the forgetting of previous tasks, it also reduces the model's capacity for adapting to the new tasks. On the contrary, neither memory-replay based methods~\cite{chaudhry2019tiny,prabhu2020gdumb,zhang2022sparsified} nor the parameter-isolation based methods put restrictions on the model parameters. Moreover, the parameter-isolation based methods can even expand the model and increase the capacity so that the model can better adapt to the new tasks~\cite{wortsman2020supermasks,rusu2016progressive,yoon2017lifelong}. For the memory-replay based methods, an increase in the buffer size directly correlates with enhanced capacity for preserving information from preceding tasks, resulting in improved performance outcomes. For the parameter-isolation based methods, more budget for model expansion is also preferred for better performance.
In a word, currently, the CGL techniques have to maintain a trade-off between the space complexity and the effectiveness in preventing forgetting. When the number of tasks becomes relatively large, how to achieve better performance with a limited memory space budget is still challenging.

\subsection{Task-free CGL}
Existing CGL techniques mostly assume that the task boundaries are available during training. For regularization based methods, task boundaries are used for matching the stored parameter importance to the corresponding tasks.
For memory replay based methods, task boundaries help allocate memory budget for different tasks. For parameter-isolation based methods, the boundaries are also crucial to determine when to expand the network. 
However, in real-world applications, data from different tasks may come in as a mixture without a clear task boundary, raising multiple challenges. First, without the given task identities, it is critical to investigate how to properly detect the distribution shift or the emergence of new patterns so that the model or memory buffer can expand. Second, since the data from a certain task may not only emerge once, even if the distribution shift is detected, the incoming `new' patterns may have already been learnt previously. In this case, if the model always treats the incoming data with distribution shift as a new task, memory space will be wasted to maintain duplicated knowledge. Therefore, how to properly manage the learnt knowledge so that the learnt knowledge is preserved effectively and efficiently is also a challenge to tackle.

\subsection{Concept Drift of Nodes in Existing Tasks}
As pointed out in CGLB~\cite{zhang2022cglb} and multiple technical works~\cite{wei2022incregnn,su2023towards}, since the inter-task edges connecting nodes from old and new tasks appear after learning on the old tasks, the affected nodes in the old tasks would experience concept drift. Currently, the only solution in the existing works~\cite{wei2022incregnn,su2023towards_seaer} is to retrain the model on these affected nodes within the updated large graph, so that the nodes after concept drift are learnt by the model. This approach assumes that all previous nodes including the affected nodes and their neighboring nodes are accessible during the learning of new tasks. However, due to reasons like privacy issues, nodes from the previous tasks may not always be accessible, which is why continual learning works typically forbid to access historical data. In light of this, how to tackle the concept drift on the affected previous nodes without accessing previous nodes is a crucial problem, especially in dense networks with a large amount of newly emerging inter-task edges.

\subsection{Exploring Task-wise Knowledge Transfer}
Currently, the focus of the CGL research is still on alleviating the forgetting problem. However, as pointed out by \cite{lopez2017gradient,9808404}, besides avoiding the negative transfer (forgetting), we also expect the task-wise transfer to be positive. In other words, the knowledge learnt from one task is expected to be transferable and can benefit the performance of other tasks. By learning transferable features, the catastrophic forgetting problem can also be overcome as a side effect since the task-wise influence becomes beneficial instead of detrimental. Works like HPNs~\cite{9808404} and RieGrace~\cite{sun2023self} have made preliminary attempts to learn transferable features via decomposing each node into basic features or self-supervised learning, respectively. But more future works are still desired not only for enhanced performance but also for better explainability of the transferable knowledge.

\subsection{Extension to Multimodal Input and Heterogeneous Graphs}
Since graph data is related to various research areas, information from different modalities may accompany the given graphs. For example, images may be attached to products along with their textual name in a co-purchasing network~\cite{cai2022multimodal}. Therefore, a natural problem is how to incorporate the information from different modalities into the continual learning process. As an early attempt mentioned in Section~\ref{sec:regularization based methods}, MSCGL~\cite{cai2022multimodal} approaches the problem by accommodating different modalities with separate GNNs. In the following, we list three promising directions for exploring the multi-modal CGL problem. First, during the learning process, information from all modalities may not be available during each task. Therefore, if the modality composition of different tasks is largely different, whether more severe forgetting would be caused and how to avoid it is a practical problem. Second, since each same object (\textit{e.g.} a graph node) is described with information of different modalities, information overlapping may exist across different modalities. Therefore, when trying to preserve the representative data, \textit{e.g.} through a memory buffer, how to avoid storing redundant information and preserve the information efficiently is also practically important. Finally,  in real-world applications, heterogeneous graphs (characterized by different types of nodes and edges) are more prevalent as they can provide a more complex and nuanced representation of relationships and entities. Therefore, how to extend CGL with multimodal information over heterogeneous graphs is also worthy of investigation.

\subsection{Beyond Graph Classification Tasks}
Currently, existing CGL works are mainly applied to sequences of classification tasks, which can be easily constructed by grouping the data according to the classes. However, there are also other important graph learning scenarios, like the graph regression tasks~\cite{jia2020residual,maron2019provably}, graph anomaly detection tasks~\cite{akoglu2015graph,ma2021comprehensive}, graph generation tasks~\cite{9920219,9661322}, \textit{etc.}. Different from classification tasks with explicit classes corresponding to data from different distributions, the boundaries between different distributions are implicit, posing challenges for task construction. It is also valuable to explore CGL over those tasks and provide more practical solutions in different scenarios.

\section{CGL and Graph Foundation Models}
Large language models (LLMs) have achieved remarkable success in serving as the foundation model for various natural language processing (NLP) downstream tasks. However, graph foundation models are still in their infancy. In this section, we will introduce how CGL techniques can be related to building effective graph foundation models.

\subsection{Facilitating in Graph Foundation Model Training}

To construct a graph foundation model, a promising approach is to design strategies to fine-tune pre-trained LLMs with tokenized graph input, so that the graph structural knowledge is aligned with the knowledge contained in language data. However, such consecutive training would inevitably trigger the catastrophic forgetting problem, since the graph data lie in significantly different domains from the language data used to pre-train the LLMs. Therefore, CGL techniques and the general continual learning techniques become indispensable in this process. 

Besides, an ideal graph foundation model, whether developed on top of pre-trained LLMs or trained from scratch, should maintain an up-to-date knowledge base by continuously integrating newly emerged data. However, this continual learning process may also incur the forgetting problem, which is also faced by current LLMs. In this case, CGL techniques should be developed to resolve this issue. Moreover, when the new graph data correspond to different domains, \textit{e.g.} a model may encounter both biological network data and molecule graph data, the cross-domain learning can also trigger the forgetting issue. In this scenario, the domain-IL CGL techniques could be developed to resolve the problem.

\subsection{A Potential Efficient Way to Develop Large Graph Models}
Unlike LLMs, which have demonstrated impressive performance in NLP tasks, the effectiveness of large models for graph data remains a question. Parameter-isolation based CGL models learn on expanding datasets that may significantly increase in size, with the model's size scaling accordingly. Therefore, it provides a potential solution to develop large graph models from scratch by incrementally expanding both the model and the dataset. On the one hand, gradually learning over the tasks instead of optimizing the model for all tasks simultaneously may reduce the optimization difficulty. On the other hand, CGL methods will only allocate new parameters when necessary~\cite{9808404}, which could help alleviate the parameter redundancy issue in large models. Besides, the existing large pre-trained graph models are typically domain specific~\cite{xia2022survey}, greatly limiting their capability. Fortunately, this challenge may potentially be tackled by leveraging the CGL techniques working under the domain-IL, which aims to train a given graph learning model consecutively across multiple different domains.

\ifCLASSOPTIONcaptionsoff
  \newpage
\fi



\bibliographystyle{IEEEtran}
\bibliography{ref}
\end{document}